\documentclass{article}

\usepackage{microtype}
\usepackage{graphicx}
\usepackage{subfigure}
\usepackage{array}
\usepackage{colortbl}
\usepackage{caption}
\usepackage{tabularray}
\usepackage{booktabs} 
\usepackage{multirow}
\usepackage{arydshln}
\usepackage{float}
\usepackage{bm}
\usepackage{algorithm}
\usepackage{algpseudocode}
\usepackage[table]{xcolor}
\definecolor{habicolor}{hsb}{0.66, 0.1, 0.95} 
\definecolor{Black}{rgb}{1,1,1}
\usepackage[normalem]{ulem}
\useunder{\uline}{\ul}{}

\usepackage{hyperref}

\usepackage[accepted]{icml2025}

\usepackage{amsmath}
\usepackage{amssymb}
\usepackage{mathtools}
\usepackage{float}
\usepackage{amsthm}
\usepackage{xspace}

\usepackage[capitalize,noabbrev]{cleveref}

\theoremstyle{plain}
\newtheorem{theorem}{Theorem}[section]

\theoremstyle{definition}

\theoremstyle{remark}

\usepackage[textsize=tiny]{todonotes}

\icmltitlerunning{State-Action Inpainting Diffuser}

\begin{document}

\twocolumn[
\icmltitle{State-Action Inpainting Diffuser for Continuous Control with Delay}



\icmlsetsymbol{equal}{*}

\begin{icmlauthorlist}
\icmlauthor{Dongqi Han}{MSRA}
\icmlauthor{William Wei Wang}{UP}
\icmlauthor{Enze Zhang}{SJTU}
\icmlauthor{Dongsheng Li}{MSRA}
\end{icmlauthorlist}

\icmlaffiliation{UP}{University of Pennsylvania}
\icmlaffiliation{SJTU}{Shanghai Jiaotong University}
\icmlaffiliation{MSRA}{Microsoft Research Asia}

\icmlcorrespondingauthor{Dongqi Han}{dongqihan@microsoft.com}
\icmlcorrespondingauthor{Dongsheng Li}{dongsli@microsoft.com}
\icmlkeywords{Machine Learning, ICML}

\vskip 0.3in
]



\printAffiliationsAndNotice{}  

\newcommand{\dongqi}[1]{{\color{blue}[[Dongqi: #1]]}}

\newcommand{\etc}{etc.}
\newcommand{\eg}{{\emph{e.g.}},\xspace}

\begin{abstract}
Signal delay poses a fundamental challenge in continuous control and reinforcement learning (RL) by introducing a temporal gap between interaction and perception. Current solutions have largely evolved along two distinct paradigms: model-free approaches which utilize state augmentation to preserve Markovian properties, and model-based methods which focus on inferring latent beliefs via dynamics modeling. In this paper, we bridge these perspectives by introducing State-Action Inpainting Diffuser (SAID), a framework that integrates the inductive bias of dynamics learning with the direct decision-making capability of policy optimization. By formulating the problem as a joint sequence inpainting task, SAID implicitly captures environmental dynamics while directly generating consistent plans, effectively operating at the intersection of model-based and model-free paradigms. Crucially, this generative formulation allows SAID to be seamlessly applied to both online and offline RL. Extensive experiments on delayed continuous control benchmarks demonstrate that SAID achieves state-of-the-art and robust performance. Our study suggests a new methodology to advance the field of RL with delay.
\end{abstract}


\section{Introduction}

Deep reinforcement learning (DRL) has evolved rapidly \citep{sutton1998reinforcement}, achieving success in virtual domains like video games \citep{vinyals2019grandmaster} and simulations \citep{haarnoja2018soft}, as well as complex real-world tasks ranging from Tokamak control \citep{degrave2022magnetic} to tuning language models \citep{schulman2017proximal, brown2020language}. However, a critical challenge in DRL is \textbf{signal delay}: where observations or actions are not instantaneously received by the decision maker. Such delays are universal in practice, arising from network latency or processing bottlenecks in autonomous navigation \citep{jafaripournimchahi2022stability}, high-frequency trading \citep{fang2021universal}, robotics \citep{abadia2021cerebellar}, and tele-medicine \citep{meng2004remote}. Even brief delays (e.g., 1 ms inference time) can be detrimental in rapidly changing environments like Tokamaks \citep{degrave2022magnetic}. These latencies significantly degrade DRL performance, necessitating urgent research into effective solutions. Furthermore, signal delay is intrinsic to biological systems. Human neural signals take approximately 150 ms \citep{gerwig2005timing} to traverse afferent (sensor-to-brain) and efferent (brain-to-muscle) pathways. This lag is significant during high-speed motor control, such as sprinting or piano playing \citep{bastian2006learning}.

Standard deep RL methods, including those designed for partially observable tasks, have been reported to experience catastrophic failure on RL with delay (RLwD), even when the delay is small. In response to this challenge, a number of studies have been proposed recently to specifically handle RLwD problems.  These methods can be categorized into two classes: \textbf{model-free} and \textbf{model-based}. 

Model-free methods \citep{revisitingstateaugmentation2021,delayedreinforcementlearning2022,reinforcementlearningwith2021,boostingreinforcementlearning2024,variationaldelayedpolicy2024} for RLwD are based on a theorem that delayed observation can be converted into a standard Markov decision process (MDP) problem by attaching previous actions into the current observation as the input for the decision-making model \citep{addressingsignaldelay2024}. Such methods are often referred to as \textbf{state-augmentation} methods, which are simple to implement, yet requiring the decision making model to learn the nature of control with delay (i.e., less inductive bias for the task).

Model-based RLwD methods \citep{agarwal2021blind,chen2021delay,learningbeliefrepresentation2021,reinforcementlearningfrom2024,directlyforecastingbelief2025} focused on estimating the true state (undelayed observation) or the \textbf{belief} of it \citep{schmidhuber1991reinforcement, han2020variational} using an explicit or latent world model (such as Dreamer \citep{hafner2019dream}). A policy model then takes the estimated true state of the belief of it as input for decision making. Meanwhile, these methods are, as one might expect, dependent on the accuracy and capacity of the employed world model.

\begin{figure}
    \centering
    \includegraphics[width=1.0\linewidth]{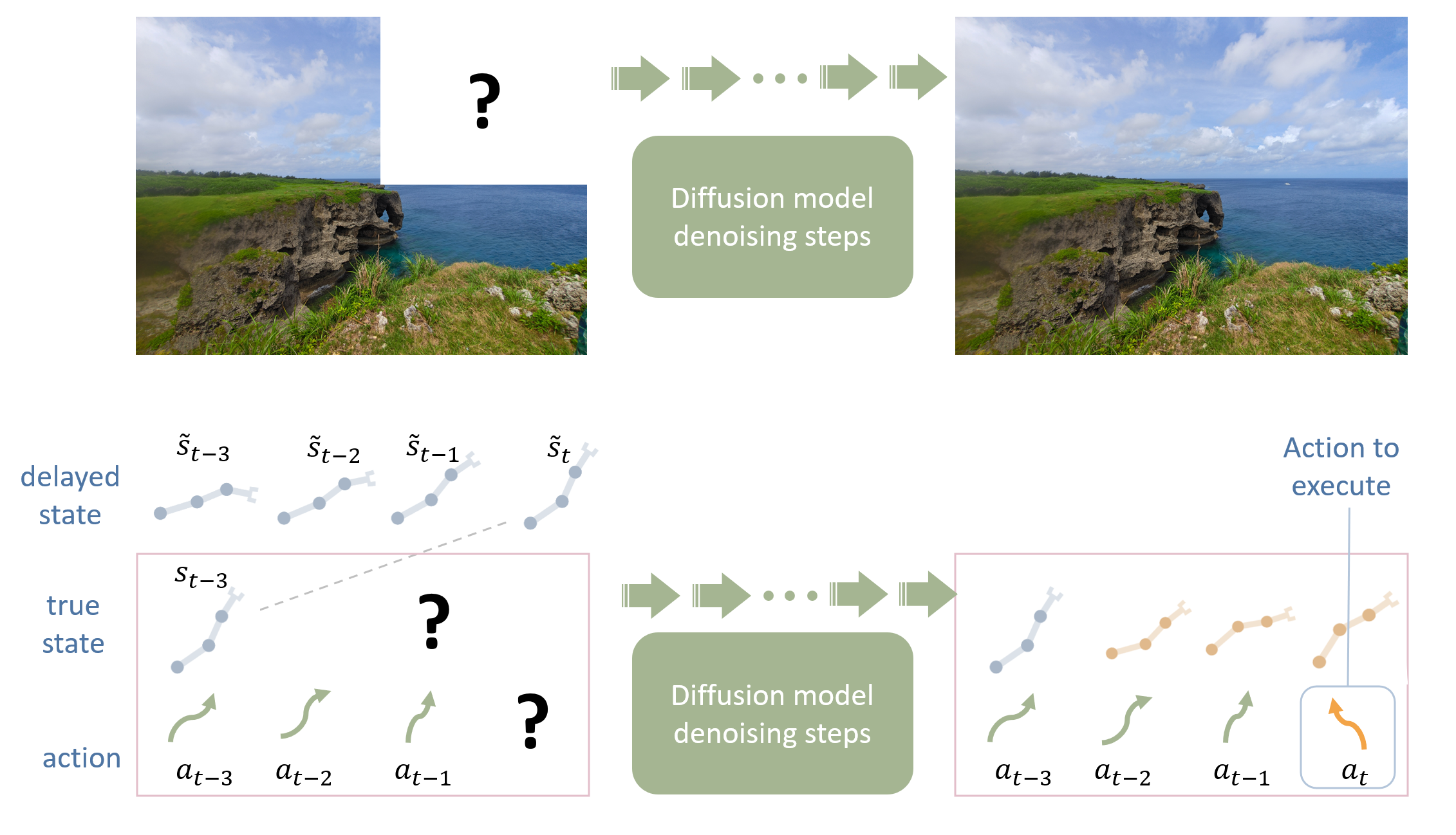}
    \caption{Illustration of our idea under case delay=3 as an example. Upper: inpainting in images. Bottom: state-action sequence inpainting for decision making with signal delay (our approach).}
    \label{fig:illustrate}
\end{figure}

Here, we propose a new approach based on diffusion planning \citep{janner2022planning}, formulating RLwD as an inpainting task (like in image generation) using diffusion models \citep{ho2020denoising}, as illustrated in Fig.~\ref{fig:illustrate}. Our approach can be considered at the intersection of model-based and model-free RLwD methods. It implicitly learns the state-transition dynamics of the environment, while also taking the current observation and previous actions as input to output the desired action directly. Our approach, referred to as \textit{State-Action Inpainting Diffuser} (SAID), integrates the advantages of model-free and model-based RLwD methods.

SAID has several key advantages. First, it performs well both with and without delay compared with different baselines (Table.~\ref{table:offline_results} \& Table.~\ref{table:online_results}). Moreover, it can be effectively applied to both online RL and offline RL (Sec.~\ref{chap:online_results} \& Sec.~\ref{chap:offline_results}), showing its versatility. Finally, it is robust to many design choices  (Sec.~\ref{chap:robustness}), indicating that the performance gain is not from the implementation tricks but the novelty of this method. In light of our results, it suggests a new category of approach rather than previous state-augmented and belief-based methods to address the signal delay problems in reinforcement learning and continuous control.

\section{Background}
\label{chap:background}

\subsection{Markov decision process and RL}

A Markov Decision Process (MDP)~\citep{bellman1957markovian} provides the formal framework underlying reinforcement learning (RL)~\cite{sutton1998reinforcement}, defined by a state space, an action space, a transition distribution, a reward function, and a discount factor. In RL, an agent interacts with the environment modeled as an MDP to learn a policy that maximizes cumulative reward. In \textit{online RL}~\citep{mnih2015human, haarnoja2018soft}, the agent learns through real-time interaction, continuously collecting new data while updating its policy. In contrast, \textit{offline RL} (or batch RL)~\citep{fujimoto2019off, levine2020offline} learns solely from a fixed dataset of previously collected trajectories, without any further environment interaction, making it particularly suitable for domains where exploration is costly, risky, or impractical.


\subsection{Delayed Observation MDP}

A \textit{Delayed observation Markov decision process} (DOMDP) \citep{chen2021delay,addressingsignaldelay2024} is defined as a special case of a POMDP by specifying its state as $\sigma=(s^{(-T)}, s^{(-T+1)}, ..., s^{(-1)}, s)$, where $T$ denotes the maximum delay and $s$ represents the state of an MDP without delay. Intuitively, $s^{(-t)}$ represents the state of the original MDP from $t$ steps prior, and the superscript $(-t)$ is used to indicate the relative timestep shift so that the Markovian property is preserved. The transition probability function of a DOMDP is defined as
\begin{align*}
\cT(\sigma', r \mid \sigma, a) = \cT_0(s', r \mid s, a)\prod_{t=1}^{T} \mathbb{I}(s'^{(-t)} = s^{(-t+1)}),
\end{align*}
where $\mathbb{I}$ denotes the indicator function and $\cT_0(s', r \mid s, a)$ denotes the transition probability of the original MDP. The term $\prod_{t=1}^{T} \mathbb{I}(s'^{(-t)} = s^{(-t+1)})$ is used to transfer $s^{(-t+1)}$ at the current step to $s^{(-t)}$ at the next step, thereby interpreting $s^{(-t)}$ as the delayed state from $t$ steps prior, without requiring the introduction of an absolute time step as in a time-dependent MDP \citep{boyan2000exact}.

The observation probability is then defined as
\begin{align*}
\cO(\tilde{s}\mid \sigma) & = \cO(\tilde{s} \mid s^{(-T)}, s^{(-T+1)}, ..., s^{(-1)}, s)  \\
& = \sum_{t=1}^T \mathcal P(t)\, \mathbb{I}(\tilde{s} = s^{(-t)}),
\end{align*}
where $\tilde{s}$ denotes the observation (i.e., the delayed state), and $\mathcal P(t)$ denotes the probability that the signal is delayed for $t$ steps. A simple case is given by $\mathcal P(t) = \mathbb{I}(t=\Delta T)$, which indicates that the signal delay is fixed as $\Delta T$ steps, implying $\tilde{s} = s^{(-\Delta T)}$. A DOMDP is thus fully defined by specifying the elements of the corresponding POMDP.

An important properties of DOMDP is described as follows (proven by \citet{addressingsignaldelay2024})

\begin{theorem}[Recovering Markovian Property by State Augmentation] \label{theorem:markovian}
By integrating historical actions \( a_{< t} \) into the observation, the process becomes an MDP with state transition probability
$
\bar{\mathcal P}(\bar s_{t+1} \mid \bar s_t, a_t),
$
where the augmented state is \( \bar s_t = (\tilde s_t, a_{t-\Delta T : t-1}) \).
\end{theorem}

\subsection{Diffusion models for decision making}

Diffusion models have recently shown impressive effectiveness in decision-making tasks, largely due to their capacity for capturing complex distributions. In contrast to classical diagonal Gaussian policies~\citep{haarnoja2019soft, schulman2017proximal, kostrikov2021offline, kumar2020conservative}, diffusion-based approaches have achieved state-of-the-art results in both online~\citep{wang2024diffusion,yang2023policy,psenka2023learning,ren2024diffusion} and offline reinforcement learning~\citep{chen2023offline, li2023hierarchical, wang2023diffusion, hansen2023idql, janner2022planning}, as well as in demonstration learning~\citep{ze20243d, chi2023diffusion}. Broadly, diffusion models are utilized in decision-making through two primary paradigms:

(1) \textit{Diffusion planner}~\cite{ajay2022conditional, janner2022planning, liang2023adaptdiffuser, lu2025what}, which models a trajectory $\tau$ capturing the current state and the following $H$ future states (or state–action pairs) over a planning horizon:
\begin{align*}
\tau = \begin{bmatrix} s_{t} , s_{t+1} , \cdots , s_{t+H-1} \end{bmatrix}
\text{ or }
\begin{bmatrix} s_{t} , s_{t+1} , \cdots , s_{t+H-1}   \\ 
a_{t} , a_{t+1} , \cdots , a_{t+H-1}
\end{bmatrix}.
\end{align*}

(2) \textit{Diffusion policy}~\cite{hansen2023idql, wang2023diffusion}, which directly models the action distribution $p(a_t \mid s_t)$ using diffusion models, and can be interpreted as trajectory modeling with horizon $H=1$, i.e., $\tau = [s_t \; a_t]$.

Recent work has further broadened the scope of diffusion planning, extending it to settings such as vision-based decision-making~\cite{chi2023diffusion} and approaches incorporating 3D visual representations~\cite{ze20243d}. A comprehensive study~\cite{lu2025what} examines crucial design factors and provides practical guidance for effective diffusion planning. Our method, SAID, falls within the class of diffusion planners.

\section{Related Work}
Recent work in RL commonly formalizes signal delay under the MDP framework, including delayed MDPs with equivalence transformations to standard MDPs \cite{revisitingstateaugmentation2021}, and modeling delayed observations as a special form of partially observable Markov decision processes (POMDPs), termed Delayed-Observation MDPs (DOMDPs) \cite{addressingsignaldelay2024}. Alongside these theoretical formulations, existing approaches can be broadly grouped into two methodological paradigms. \textbf{Augmentation-based approaches} optimize policies in an augmented MDP by concatenating delay-window histories to the policy input. For instance, DIDA \cite{delayedreinforcementlearning2022} learns delayed policies in the augmented MDP by imitating delay-free expert behaviors; DCAC performs delay-corrected actor--critic updates via off-policy multi-step value estimation under stochastic delays \cite{reinforcementlearningwith2021}; AD-RL bootstraps long-delay learning using auxiliary short-delay tasks \cite{boostingreinforcementlearning2024}; and VDPO improves optimization efficiency by casting delayed RL in a variational inference framework \cite{variationaldelayedpolicy2024}. However, augmentation-based methods operate over increasingly high-dimensional augmented state spaces as the delay horizon grows, which can lead to scalability issues and degraded sample efficiency.
\textbf{Belief-/prediction-based approaches} instead infer the current latent state (belief) from delayed observations and optimize policies based on the inferred belief, avoiding delay-window expansion; their performance is therefore tightly coupled to belief estimation accuracy. 
Representative early works explicitly model delay-aware dynamics and belief estimation, including EMQL \cite{agarwal2021blind}, which selects actions by maximizing expected Q-values under the inferred belief, DATS in delay-aware model-based RL \cite{chen2021delay} and attention-based belief prediction as in D-SAC \cite{learningbeliefrepresentation2021}, while world-model-based approaches such as D-Dreamer reduce delayed POMDPs to (belief) MDPs and enable visual delayed continuous control \cite{reinforcementlearningfrom2024}. A challenge for forecasting-based belief estimation is error accumulation under recursive multi-step prediction, which can lead to compounding errors; DFBT directly forecasts beliefs to mitigate such compounding effects compared to step-by-step prediction \cite{directlyforecastingbelief2025}.

\section{Methods}

\subsection{Nomenclature}
To make the symbols and variable names clear, here we details the essential quantities as follows. The environment step in an episode is denoted as $t$, which is used for the subscript for the state $s_t=s$ at step $t$. In a DOMDP with delay step $\Delta T$, the observation, namely the delayed state, is denoted by $\tilde{s}_t$. which is equal to $s_{t-\Delta T}$. Action and reward are denoted by $a_t$ and $r_t$.

\subsection{State-action inpainting for DOMDP}
 




In the context of a Delayed-Observation MDP (DOMDP), the objective is to recover the underlying Markovian dynamics from delayed signals. Theoretically, since the delayed state $\tilde{s}_{t}$ is simply a time-shifted version of the true state $s_t$, it should be possible to infer the current true state by applying the transition dynamics recursively $\Delta T$ times. Conventional approaches often attempt to learn a forward dynamics model and perform multi-step forecasting to bridge the delay gap \citep{chen2021delay}. However, while this approach is theoretically sound, it frequently results in catastrophic failure in practice, particularly as the delay horizon increases \citep{addressingsignaldelay2024}.

The fundamental limitation of explicit belief estimation lies in the nature of the regression task involved in continuous control. Unlike classification, where decision boundaries can be robust to minor perturbations, continuous state regression is highly sensitive to precision. As demonstrated empirically in our motivation verification (Section~\ref{chap:motivation}), autoregressive prediction suffers severely from the ``compounding error'' phenomenon. When a model feeds its own prediction back as input for the next step, small approximation errors at $t=0$ accumulate and amplify exponentially over the planning horizon $H$. This leads to a significant distributional drift, where the predicted trajectory diverges rapidly from the ground truth dynamics (see Figure~\ref{fig:motivation}), rendering the estimated current state unreliable for policy input.

To alleviate this problem, we propose a paradigm shift from recursive forecasting to generative inpainting, grounded in two key insights. First, we propose to generate the undelayed observation using a diffusion model rather than an autoregressive model. Diffusion models generate the entire sequence holistically in a non-autoregressive manner, thereby avoiding the sequential accumulation of single-step errors \citep{janner2022planning}. Second, we propose to jointly generate both state and action sequences. Since the transition dynamics in a DOMDP are coupled with the action history, modeling the joint distribution $p(\tau)$ of state-action pairs allows the model to capture the complex dependencies required to recover the underlying dynamics more accurately than state-only prediction.

\begin{figure}
    \centering
    \includegraphics[width=1.0\linewidth]{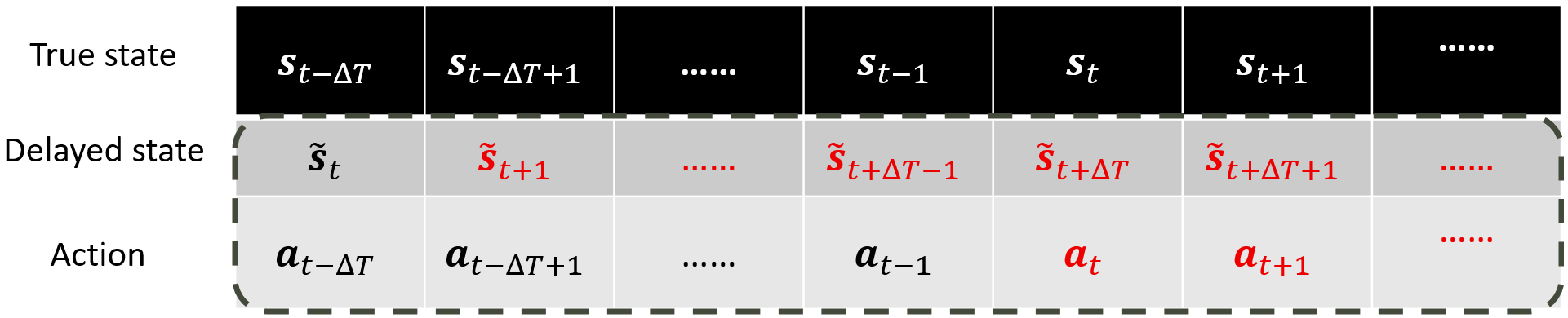}
    \caption{The being modeled state-action sequence (in the dashed rectangle) of our State-Action Inpainting Diffuser at environment timestep $t$ with delay $\Delta t$. Black color indicates input conditions and red color indicates the inpainting values.}
    \label{fig:method}
    \vspace{-2mm}
\end{figure}

Building on these insights, we introduce the \textbf{State-Action Inpainting Diffuser (SAID)}. We formulate the delay problem as a sequence inpainting task, analogous to filling in missing regions in an image. We define a trajectory of horizon $H$ consisting of state-action pairs (Fig.~\ref{fig:method}). In the DOMDP setting, the history of delayed observations and executed actions is known (the "condition"), while the current true state and optimal future actions are unknown (the "mask"). During inference, we condition the diffusion process on the available delayed history and allow the model to ``inpaint'' the missing state and actions simultaneously (red-colored variables in Fig.~\ref{fig:method}). This approach effectively recovers the Markovian state $s_t$ and generates a consistent plan without explicit state stacking or recursive estimation.

We adopt the Diffuser architecture \citep{janner2022planning} but replaced the original U-Net \citep{ronneberger2015u} by the Diffusion Transformer (DiT) architecture \citep{peebles2023scalable} for the Denoising Diffusion Implicit Models (DDIM) \citep{song2020denoising} planner. At each decision step, we employ Monte-Carlo Sampling with Selection (MCSS) \citep{lu2025what}---at each decision step, the planner samples multiple candidates of state-action trajectories, and a critic function is used to select the highest-valued trajectory to make the final decision. Details are deferred to Appendix.~\ref{appendix:implementation}.

\subsection{Practical algorithm}





The State-Action Inpainting Diffuser (SAID) framework\footnote{Source code is available in Supplementary Materials.} fundamentally relies on a dataset of state-action sequences to train a diffusion planner that can recover Markovian dynamics from delayed observations. In the offline reinforcement learning setting described in Alg.~\ref{alg:offline_rl}, the method simply utilizes the provided static dataset, preprocessing it to align delayed observations with their corresponding actions. Conversely, for the online reinforcement learning context outlined in Alg.~\ref{alg:online_rl}, the system actively collects this training data by employing an existing delay-aware RL algorithm to interact with the environment and populate a replay buffer. 

A primary distinction between the two approaches lies in value function learning; the online method utilizes the value function maintained by the existing online agent, whereas the offline method must explicitly train a value function from the offline dataset, such as Implicit Q-Learning \citep{kostrikov2021offline}, which we adopted in this study. Despite these differences in data acquisition and training, the remaining inference mechanism is shared across both paradigms, where the diffusion model generates candidate trajectories and the value function selects the optimal action for execution. More details of implementation are deferred to Appendix~\ref{appendix:implementation}.

\subsection{Handling initial steps}

\begin{figure}[h]
    \centering
    \includegraphics[width=1.0\linewidth]{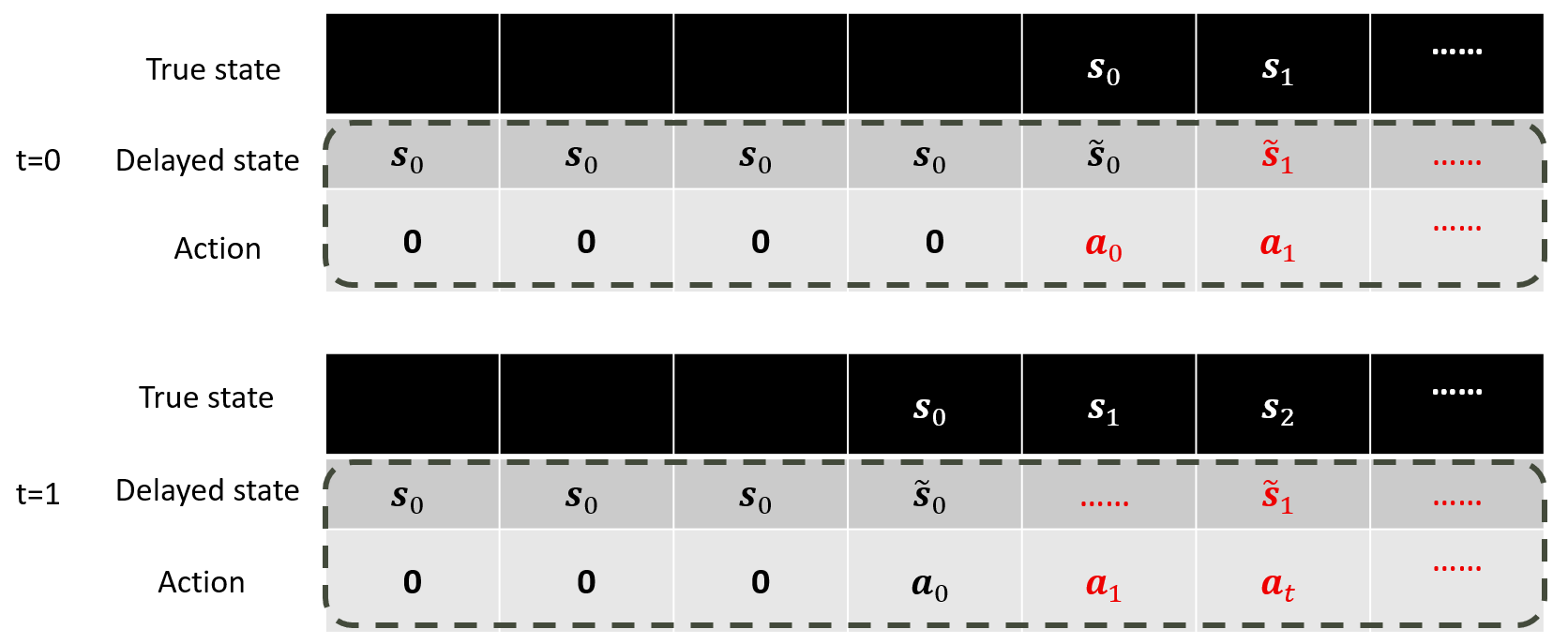}
    \caption{Handling episode start by padding initial actions and states, using $t$=0 and $t$=1 as examples. The symbols are the same as those in Figure~\ref{fig:method}.}
    \label{fig:init_steps}
    \vspace{-2mm}
\end{figure}

Fig.~\ref{fig:init_steps} illustrates the method's strategy for handling the start of an episode, where a complete history of delayed observations is not yet available. It depicts the use of padding (filling missing historical values with zeros) for the initial actions and delayed states at early timesteps (e.g., $t=0$ and $t=1$) to allow the diffusion model to generate the true state and action sequences from the very beginning of a task.

\section{Motivation Verification}
\label{chap:motivation}
\begin{figure}[h]
    \centering
    \includegraphics[width=1.0\linewidth]{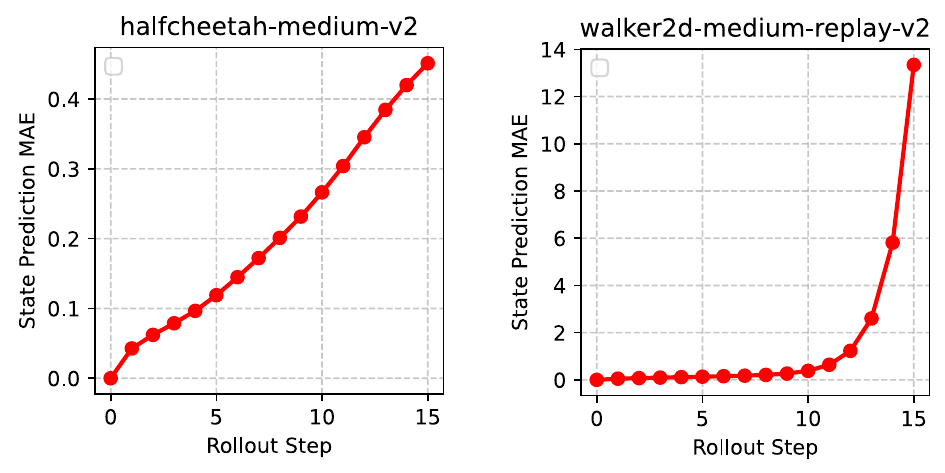}
    \caption{The ``compounding error'' phenomenon of autoregressive state-transition models. See Fig.~\ref{fig:compound} for more examples.}
    \vspace{-2mm}
    \label{fig:motivation}
\end{figure}
To empirically verify the hypothesis that prediction errors accumulate in autoregressive state-transition models (often referred to as the ``compounding error'' phenomenon), we conducted a toy experiment using D4RL MuJoCo locomotion datasets \citep{fu2020d4rl}. We established a baseline state-transition model parameterized by a multi-layer perceptron (MLP), which approximates the state transition function $\hat{s}_{t+1} = s_t + f_\theta(s_t, a_t)$. The model was trained via standard supervised learning on single-step transitions, minimizing the reconstruction loss given ground-truth current states (i.e., teacher forcing). During evaluation, we tested the model in an open-loop autoregressive setting over a fixed horizon $H$. Crucially, the model's own predicted state $\hat{s}_t$---rather than the ground truth---was fed back as input for the subsequent prediction step. We then recorded the mean absolute error (MAE) between the predicted and ground-truth states at each timestep $t$ to quantify the divergence of the generated trajectory over time (Fig.~\ref{fig:motivation}). This analysis quantifies the severity of distributional drift in autoregressive state prediction, thereby providing empirical justification for the adoption of our diffusion model-based generation methods.

\section{Experiment Results}

Following previous studies on RLwD \citep{chen2021delay, addressingsignaldelay2024, variationaldelayedpolicy2024, zhan2025adapting}, we examined SAID on a set of robot locomotion tasks with delay, since these continuous control tasks are largely affected by signal delay in reality. SAID can be applied to both online and offline RL tasks, which will be discussed in the next two sections. We tested delay steps $=0,4,8,16$. For each delay step, the actual physical time is 8 milliseconds for Hopper and Walker and 50 milliseconds for HalfCheetah and Ant robots.

\subsection{Online RL}
\label{chap:online_results}

\textbf{Benchmark tasks.}  Following standard practice in RLwD researches \citep{addressingsignaldelay2024, variationaldelayedpolicy2024}, we examined SAID on the latest version of continuous control tasks from Gymnasium \citep{towers2024gymnasium}: Halfcheetah-v5, Ant-v5, Walker2d-v5, and Hopper-v5.

\textbf{Baseline methods.}
We compare SAID against several representative online RL baselines for delayed control.
Vanilla SAC \citep{haarnoja2018soft} serves as a standard RL baseline that does not explicitly account for observation delays.
We further evaluate delay-aware approaches, including state augmentation (SA-MLP) and delay-reconciled critic (DR-Critic) \citep{addressingsignaldelay2024}. 
We additionally include variational delayed policy optimization (VDPO) \citep{variationaldelayedpolicy2024}, which learns delayed policies by imitating a policy trained in a non-delayed environment.

\begin{figure}[h]
    \centering
    \includegraphics[width=1.0\linewidth]{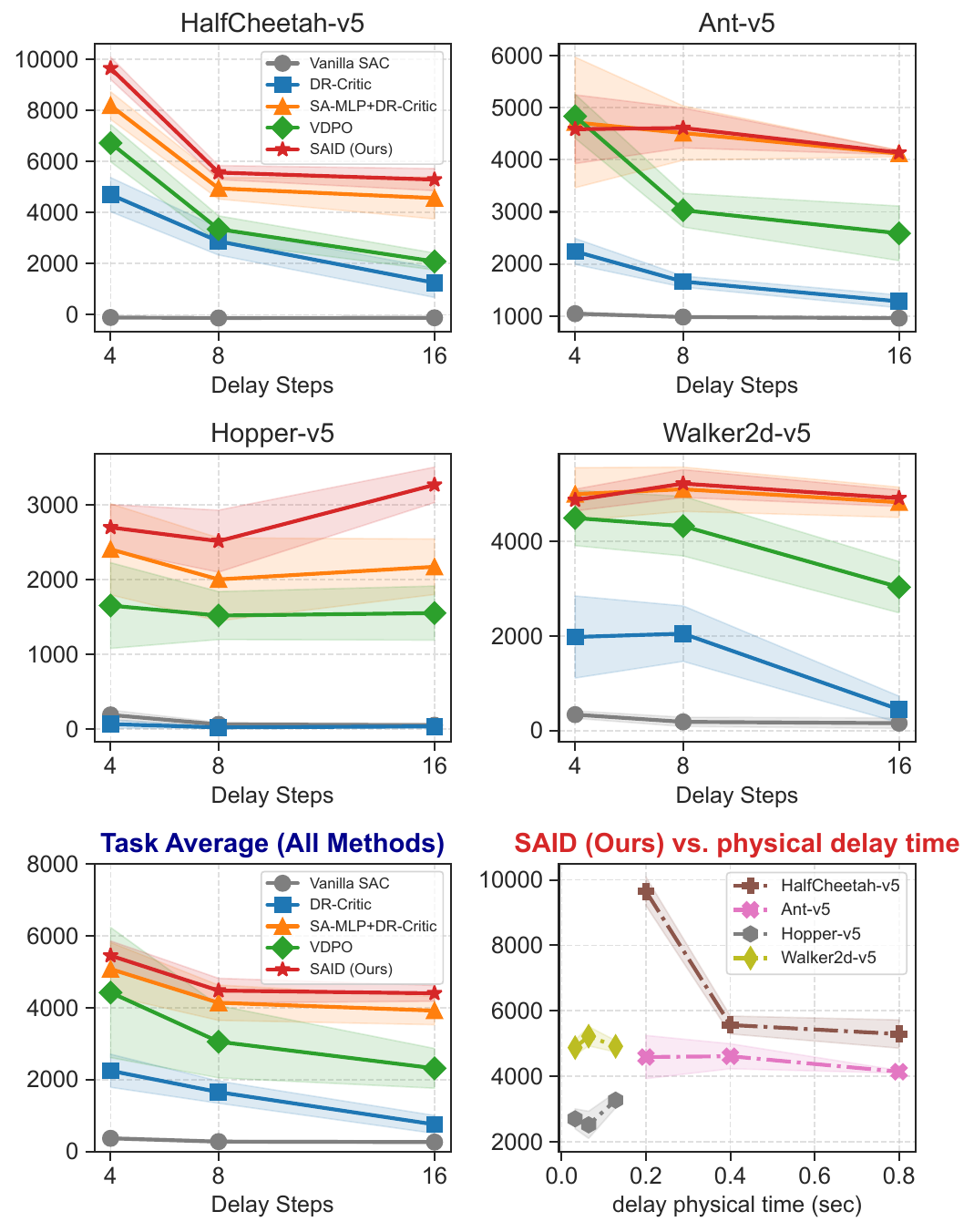}
    \caption{Performance (mean episodic return) for online RL with delay. One may notice that the experimental result of Hopper at delay=16 is higher than that at delay=8. This is also reported in Table 10 of the reference paper \cite{directlyforecastingbelief2025}.}
    \label{fig:online_perf}
    \vspace{-2mm}
\end{figure}

Fig.~\ref{fig:online_perf} and Table.~\ref{table:online_results} demonstrate that SAID consistently outperforms competing baseline methods across online reinforcement learning benchmarks subject to signal delay. In the evaluation of four MuJoCo locomotion environments, SAID achieves the highest mean episodic returns across all tested delay durations ($4$, $8$, and $16$ steps), exhibiting superior stability compared to baselines. The ``Task Average'' plot further substantiates this superiority, showing that while standard methods such as Vanilla SAC suffer catastrophic failure and other baselines degrade significantly as latency increases, SAID maintains robust performance.

\begin{table}[tbh]
\centering
\resizebox{1.0\linewidth}{!}{
\setlength{\tabcolsep}{2.5pt}
\renewcommand{\arraystretch}{1.15}

\begin{tabular}{llcccc}
\toprule
\multirow{2}{*}{Task} & \multirow{2}{*}{Method} & \multicolumn{4}{c}{Delay steps} \\
\cmidrule(lr){3-6}
 & & 0 & 4 & 8 & 16 \\
\midrule

\multirow{5}{*}{HC-v5}
& SAC
& $12450\!\pm\!695$ & $-109\!\pm\!15$ & $-129\!\pm\!18$ & $-121\!\pm\!11$ \\
& DR
& \textemdash & $4702\!\pm\!668$ & $2873\!\pm\!537$ & $1249\!\pm\!574$ \\
& SADR
& \textemdash & $8190\!\pm\!543$ & $4945\!\pm\!417$ & $4565\!\pm\!816$ \\
& VDPO
& \textemdash & $6720\!\pm\!733$ & $3350\!\pm\!520$ & $2085\!\pm\!332$ \\
& SAID
& $\mathbf{14845\!\pm\!367}$ & $\mathbf{9648\!\pm\!436}$ & $\mathbf{5565\!\pm\!274}$ & $\mathbf{5289\!\pm\!432}$ \\
\midrule

\multirow{5}{*}{Ant-v5}
& SAC
& $4553\!\pm\!2017$ & $1047\!\pm\!30$ & $982\!\pm\!13$ & $961\!\pm\!6$ \\
& DR
& \textemdash & $2241\!\pm\!249$ & $1662\!\pm\!107$ & $1282\!\pm\!123$ \\
& SADR
& \textemdash & $\mathbf{4716\!\pm\!1253}$ & $4512\!\pm\!521$ & $4118\!\pm\!62$ \\
& VDPO
& \textemdash & $4834\!\pm\!422$ & $3032\!\pm\!324$ & $2588\!\pm\!524$ \\
& SAID
& $\mathbf{5953\!\pm\!84}$ & $4586\!\pm\!660$ & $\mathbf{4610\!\pm\!383}$ & $\mathbf{4137\!\pm\!41}$ \\
\midrule

\multirow{5}{*}{Hop.-v5}
& SAC
& $2354\!\pm\!645$ & $188\!\pm\!65$ & $61\!\pm\!27$ & $50\!\pm\!27$ \\
& DR
& \textemdash & $65\!\pm\!58$ & $21\!\pm\!10$ & $34\!\pm\!21$ \\
& SADR
& \textemdash & $2405\!\pm\!610$ & $2002\!\pm\!556$ & $2171\!\pm\!372$ \\
& VDPO
& \textemdash & $1652\!\pm\!574$ & $1518\!\pm\!321$ & $1551\!\pm\!362$ \\
& SAID
& $\mathbf{3170\!\pm\!283}$ & $\mathbf{2698\!\pm\!305}$ & $\mathbf{2514\!\pm\!415}$ & $\mathbf{3268\!\pm\!237}$ \\
\midrule

\multirow{5}{*}{Walk.-v5}
& SAC
& $4910\!\pm\!501$ & $338\!\pm\!79$ & $185\!\pm\!98$ & $157\!\pm\!111$ \\
& DR
& \textemdash & $1979\!\pm\!868$ & $2050\!\pm\!587$ & $447\!\pm\!281$ \\
& SADR
& \textemdash & $\mathbf{4999\!\pm\!559}$ & $5102\!\pm\!468$ & $4827\!\pm\!320$ \\
& VDPO
& \textemdash & $4494\!\pm\!584$ & $4323\!\pm\!630$ & $3032\!\pm\!542$ \\
& SAID
& $\mathbf{5158\!\pm\!360}$ & $4874\!\pm\!230$ & $\mathbf{5222\!\pm\!295}$ & $\mathbf{4913\!\pm\!175}$ \\
\midrule

\multirow{5}{*}{Task Avg.}
& SAC
& $6067\!\pm\!965$ & $366\!\pm\!47$ & $275\!\pm\!39$ & $262\!\pm\!39$ \\
& DR
& \textemdash & $2247\!\pm\!461$ & $1652\!\pm\!310$ & $753\!\pm\!250$ \\
& SADR
& \textemdash & $5078\!\pm\!741$ & $4140\!\pm\!491$ & $3920\!\pm\!393$ \\
& VDPO
& \textemdash & $4425\!\pm\!1812$ & $3056\!\pm\!1007$ & $2314\!\pm\!553$ \\
& SAID
& $\mathbf{7282\!\pm\!274}$ & $\mathbf{5452\!\pm\!408}$ & $\mathbf{4478\!\pm\!342}$ & $\mathbf{4402\!\pm\!221}$ \\
\bottomrule
\end{tabular}
}
\caption{Results across methods under different delays (mean $\pm$ standard error across 4 training random seeds). Method abbreviations: SAC (Vanilla SAC), DR (DR-Critic), SADR (SA-MLP+DR-Critic).
For delay=0, DR-Critic and SA-MLP+DR-Critic reduce to Vanilla SAC; hence we report a single SAC(0) baseline and mark other methods with \textemdash. The results of SAID are obtained from 4 random seeds with 200 evaluation episodes per seed. Bold indicates the best mean within each task and delay.}
\label{table:online_results}
\vspace{-2mm}
\end{table}

\subsection{Offline RL}

\label{chap:offline_results}

\textbf{Benchmark tasks.} We used the locomotion tasks from the standard D4RL \citep{fu2020d4rl} benchmark, which contains 3 environments (halfcheetah, hopper, walker2d), each with 3 kinds of data (medium, medium-expert, medium-replay). 

\textbf{Baseline methods.} As we are not aware of any published work adapting offline RL with online delays\footnote{While we acknowledge the existence of \citet{zhan2025adapting} on arXiv, we exclude it from our primary comparison. Despite our requests, the authors did not provide code or sufficient implementation details to replicate their results. Following standard scientific practice, we restrict our comparison to peer-reviewed or reproducible methods to ensure a fair evaluation}, we compare our method to the SOTA diffusion-based method---diffusion Q-learning \citep{wang2023diffusion} with a modification to add state-augmentation.  For a fair comparison, we also performed a grid search for the policy temperature of DQL+SA and report the best results. We also report the performance (delay=0) of representative, SOTA offline RL methods from various categories of algorithms for reference, such as CQL \citep{kumar2020conservative}, ReBRAC \citep{tarasov2023revisiting}, Diffuser \citep{janner2022planning}, DV \citep{lu2025what}. For ensemble-based methods (e.g. \citet{an2021uncertainty}, \citet{yang2022rorl}), we omit the comparison due to the orthogonality between our contributions.

\begin{figure}
    \centering
    \includegraphics[width=0.8\linewidth]{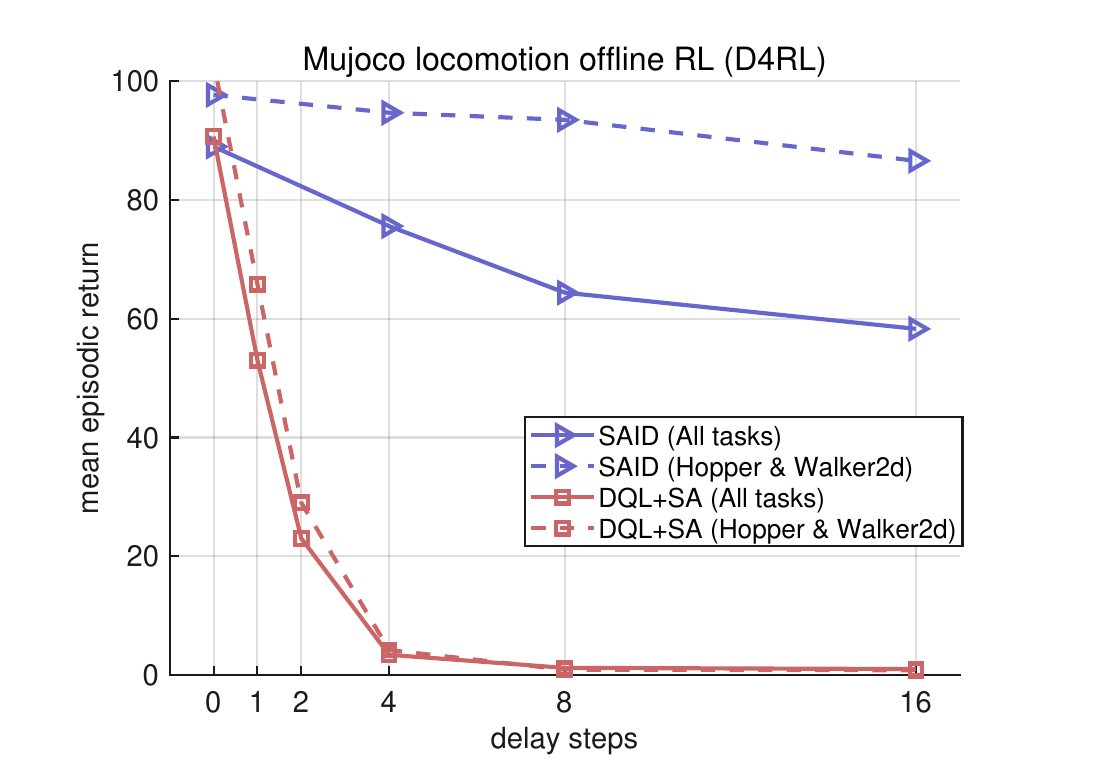}
    \caption{Offline RL performance, compared with Diffusion Q-Learning (DQL) with state augmentation (SA).}
    \vspace{-4mm}
    \label{fig:compare_with_dql}
\end{figure}

We present the results in and Table~\ref{table:offline_results}, SAID exhibits significantly superior robustness and performance in offline reinforcement learning tasks with signal delay compared to the baseline Diffusion Q-Learning with State Augmentation (DQL+SA). Fig.~\ref{fig:compare_with_dql} visually demonstrates that whereas DQL+SA suffers severely when the delay increases to 4 steps. SAID maintains high episodic returns with only a gradual decline, even extending to 16 delay steps (Table~\ref{table:offline_results}). These findings strongly suggest that the inpainting-based formulation of SAID effectively mitigates the compounding errors inherent in state-augmentation methods for delayed offline control.

Also, for the case of delay=0, SAID shows comparable performance to DQL, significantly outperforming SOTA diffusion planning method DV \cite{lu2025what}. Since the performance without delay is not the focus of this study, the result of SAID at delay=0 is provided mainly for reference to the performance degradation with delay (Fig.~\ref{fig:compare_with_dql}).

\begin{table*}
\centering
\resizebox{1.0\linewidth}{!}{
\SetTblrInner{rowsep=0pt}
\begin{tblr}{
  colspec = {l *{14}{c}},
  cells = {c},
  cell{1}{2}  = {c=3}{},
  cell{1}{5}  = {c=2}{},
  cell{1}{7}  = {c=6}{},
  cell{1}{13} = {c=3}{},
  cell{2}{9}  = {c=4}{},
  cell{2}{13} = {c=3}{},
  cell{3}{2}  = {c=7}{},
  vline{1,2,5,7,9,13,16} = {1-15}{},
  hline{1,2,3,4,13,15} = {-}{},
}
\textbf{Category} & Non-diffusion & & & Diffusion Policies & & Diffusion Planners & & & & & & Baseline w/ delay & & \\
\textbf{Method}   & BC & CQL & ReBRAC & DQL & DQL* & Diffuser & DV & SAID (ours) & & & & DQL+SA & & \\
\textbf{Delay}    & \SetCell{c=7}{No delay} & & & & & & &0 & 4 & 8 & 16 & 4 & 8 & 16 \\

halfcheetah-medium-expert-v2 & 35.8 & 62.4 & 108.1 & 96.8 & 95.9 & 88.9 & 92.7 & 106.2 & 32.8 & 4.5 & 1.9 & 2.9 & 2.1 & 1.4 \\
halfcheetah-medium-replay-v2 & 38.4 & 46.2 & 51.2  & 47.8 & 48.9 & 37.7 & 45.8 & 50.2  & 34.3 & 8.3 & 2.1 & 0.2 & 1.4 & 1.3 \\
halfcheetah-medium-v2        & 36.1 & 44.4 & 66.4  & 51.1 & 52.6 & 42.8 & 50.4 & 58.5  & 45.2 & 5.8 & 1.5 & 2.4 & 2   & 1.3 \\

walker2d-medium-expert-v2    & 6.4  & 111  & 104.6 & 110.1& 112  & 106.9& 109.2& 111.1 & 111.1& 110.8& 106.7& 8.1 & -0.3& -0.1 \\
walker2d-medium-replay-v2    & 11.8 & 26.7 & 89.4  & 95.5 & 99.9 & 70.6 & 85   & 95.5  & 91.1 & 85.6 & 45.2 & 7.1 & -0.1& -0.2 \\
walker2d-medium-v2           & 6.6  & 79.2 & 92.7  & 87   & 91.2 & 79.6 & 82.8 & 84.4  & 84.8 & 84.1 & 80.3 & -0.1& -0.1& -0.1 \\

hopper-medium-expert-v2      & 111.9& 98.7 & 109.8 & 111.1& 111.1& 103.3& 110.1& 112.3 & 111.4& 111.2& 110.3& 4.2 & 1.9 & 1 \\
hopper-medium-replay-v2      & 11.3 & 48.6 & 97.2  & 101.3& 102.9& 93.6 & 91.9 & 98.3  & 87.5 & 85.6 & 90.3 & 2.7 & 1.8 & 2.1 \\
hopper-medium-v2             & 29   & 58   & 91.1  & 90.5 & 102  & 74.3 & 83.6 & 84.6  & 82   & 83.5 & 86.6 & 3.1 & 2   & 2 \\

\textbf{All Tasks Average}        & \textbf{31.9} & \textbf{63.9} & \textbf{90.1} & \textbf{87.9} & \textbf{90.7} & \textbf{77.5} & \textbf{83.5} & \textbf{89.0} & \textbf{75.6} & \textbf{64.4} & \textbf{58.3} & \textbf{3.4} & \textbf{1.2} & \textbf{1.0} \\
\textbf{Walker \& Hopper Average} & \textbf{29.5} & \textbf{70.4} & \textbf{97.5} & \textbf{99.3} & \textbf{103.2}& \textbf{88.1} & \textbf{93.8} & \textbf{97.7} & \textbf{94.6} & \textbf{93.5} & \textbf{86.6} & \textbf{4.2} & \textbf{0.9} & \textbf{0.8} \\
\end{tblr}
}
\caption{Offline RL results on D4RL MuJoCo locomotion tasks, the performance of SAID is averaged over 4 random model seeds with each evaluated over 200 episodes. Here we omit the confidence interval since the performance of SAID for offline RL is less sensitive to random seed, see Sec.~\ref{chap:confidence} for more discussion. DQL* is our replicated results using the same code base as SAID.}
\label{table:offline_results}
\end{table*}

\subsection{Random seed matters in online learning, not offline}
\label{chap:confidence}

To investigate the robustness of our method, SAID, across different experimental conditions, we analyzed the impact of random seed initialization on performance stability. We compared the normalized standard deviation (STD) of the returns between Online RL and Offline RL settings across different delay steps.

\begin{figure}[h]
    \centering
    \includegraphics[width=0.9\linewidth]{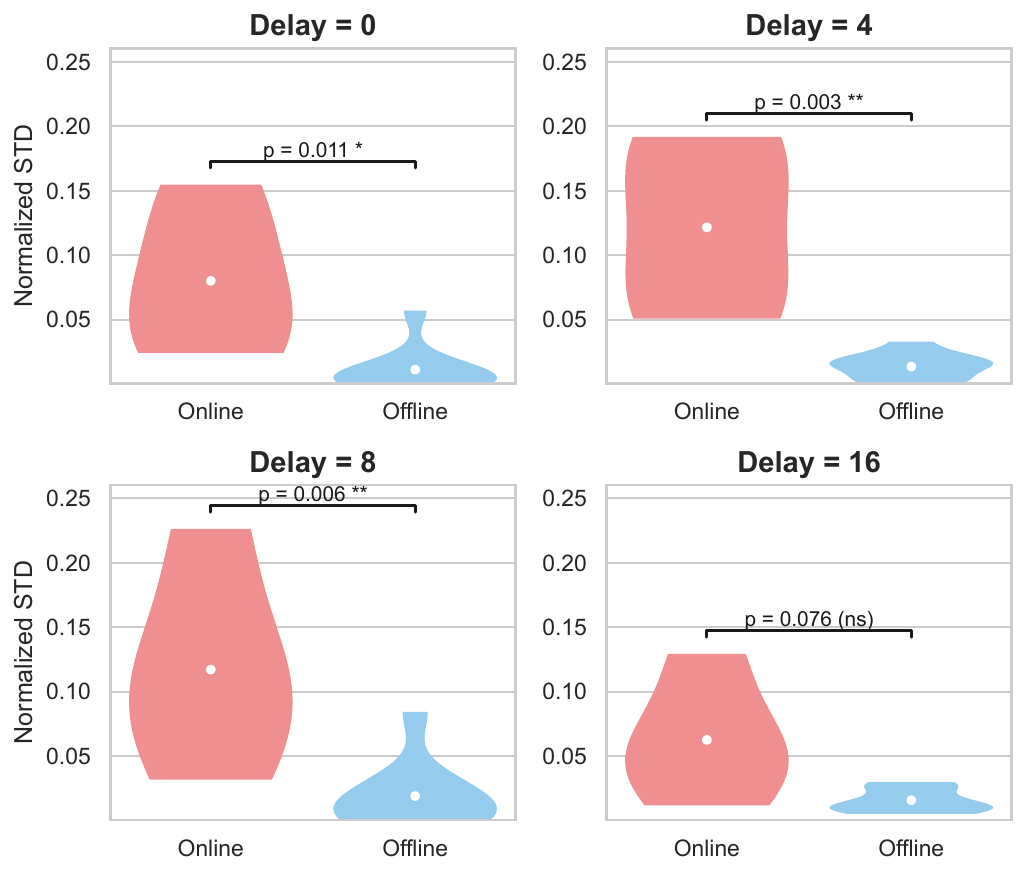}
    \caption{Violin plot of relative uncertainly (STD) of Online RL vs. offline RL among different same training seeds (averaged over 200 evaluation episdoes). The white point indicates the data mean. Mann-Whitney U-test was conducted and it showed a significantly higher seed-uncertainty of online RL than offline RL.}
    \label{fig:std_comparison_seed}
    \vspace{-2mm}
\end{figure}

Fig.~\ref{fig:std_comparison_seed} presents the uncertainty arising from different training seeds. We observed that Online RL exhibits significantly higher variance across training seeds compared to Offline RL. Specifically, Mann-Whitney U-tests indicated statistically significant differences in seed-uncertainty for lower delays. This suggests that the performance of Online RL is highly sensitive to the initial random seed (i.e., "lucky seeds"), whereas Offline RL maintains consistent performance regardless of initialization. The difference became less significant at extreme delays, likely due to the generally increased difficulty of the task for both paradigms.

\begin{figure}[h]
    \centering
    \includegraphics[width=0.9\linewidth]{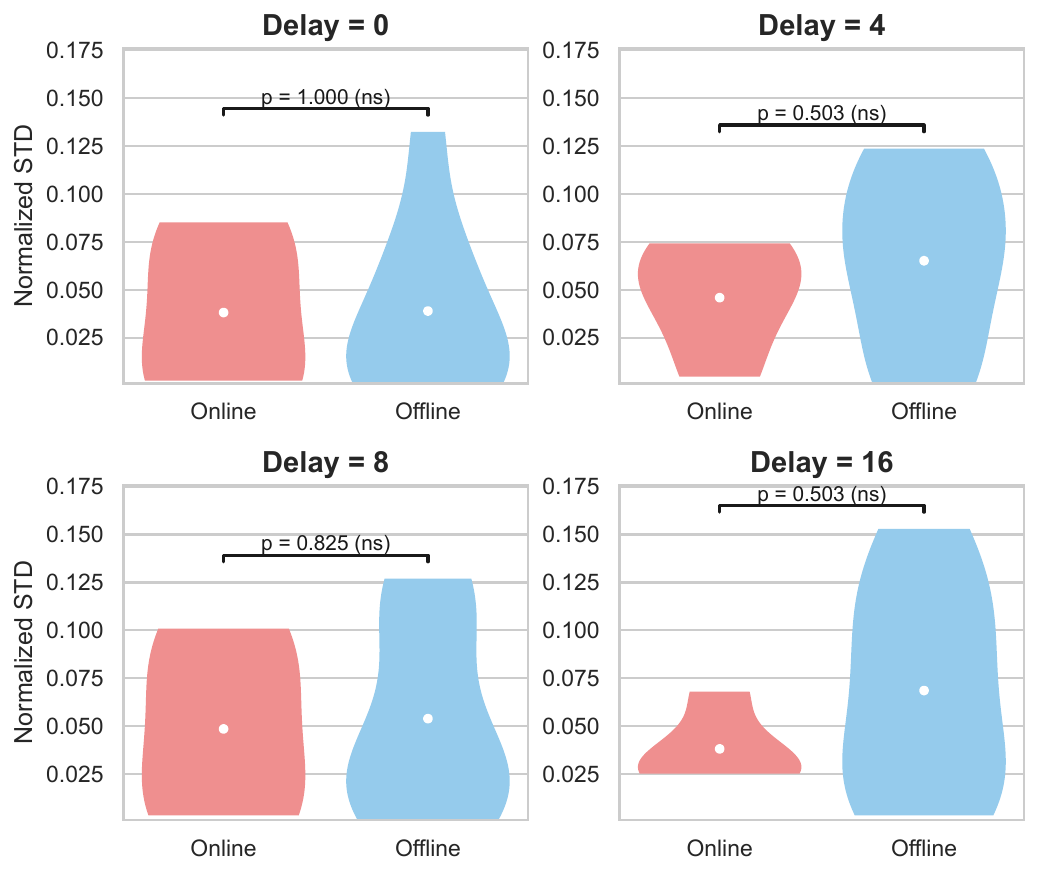}
    \caption{Violin plot of relative uncertainly (STD) of Online RL vs. offline RL among different evaluation episodes (averaged over 4 random training seeds).  The white point indicates the data mean. Mann-Whitney U-test was conducted but no significant difference was observed. }
    \label{fig:std_comparison_episodes}
    \vspace{-3mm}
\end{figure}

In contrast, Fig.~\ref{fig:std_comparison_episodes} illustrates the uncertainty across different evaluation episodes for a fixed trained policy. Here, we found no significant difference between the variances of Online and Offline RL. This indicates that once the models are trained, the stochasticity of the policy execution is comparable between the two approaches.

In conclusion, while SAID proves effective in both settings, its sensitivity to random seeds is much lower in offline RL. Therefore we report only the mean performance in Table~\ref{chap:offline_results}.

\subsection{Robustness analysis}
\label{chap:robustness}
\begin{figure}
    \centering
    \includegraphics[width=1.0\linewidth]{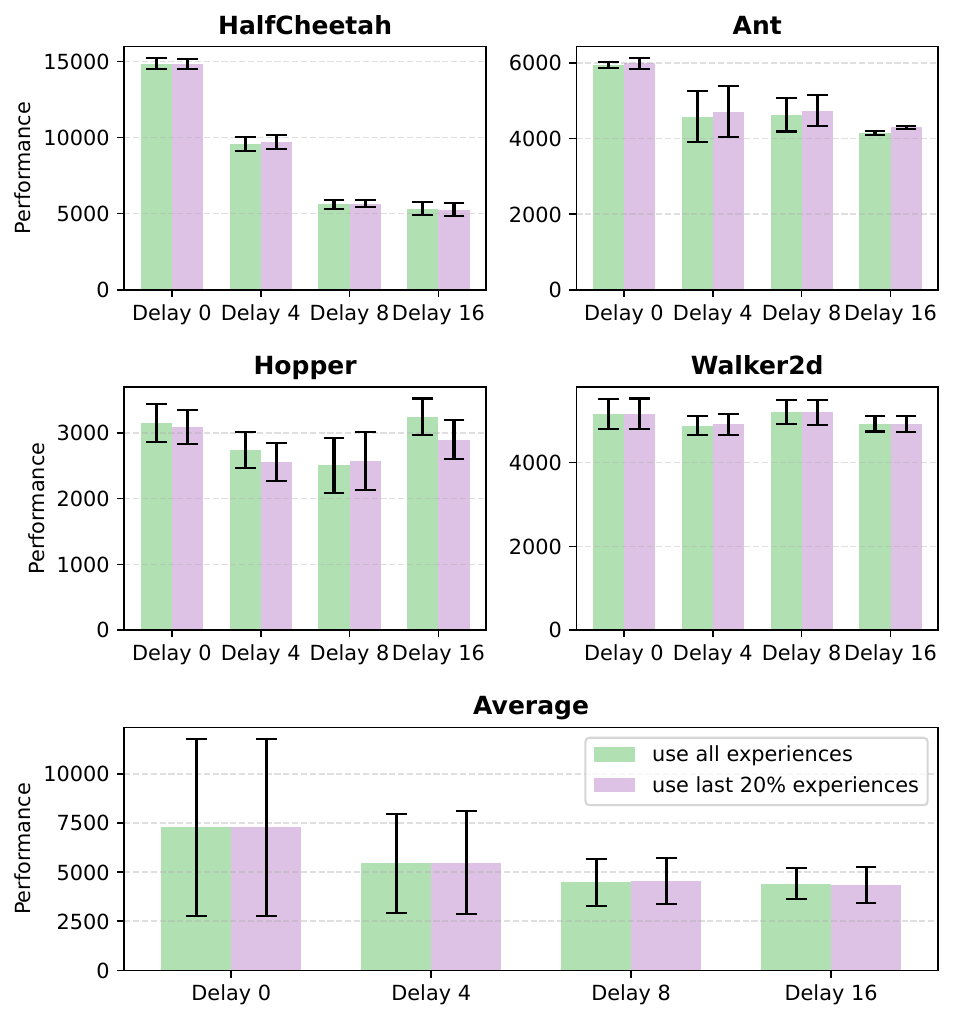}
    \caption{Examining SAID's performance for online RLwD tasks when using all online experiences (5M steps) or using only the last 20\% experiences (1M steps). Data are Mean $\pm$ S.E.M.}
    \label{fig:last}
    \vspace{-4mm}
\end{figure}
We also conducted experiments to further investigate how robust SAID is to different design choices (ablation studies).

Firstly, in Fig.~\ref{fig:last}, we observed that SAID maintains robust performance even no matter we use the full data from online experience (Algorithm~\ref{alg:online_rl}) or only the most recent 20\% of experiences (1 million steps, which contain relatively better, converged policies). Across all tested environments and delays, the average returns remain comparable, with overlapping error bars indicating no statistically significant performance difference.

\begin{figure}
    \centering
    \includegraphics[width=0.9\linewidth]{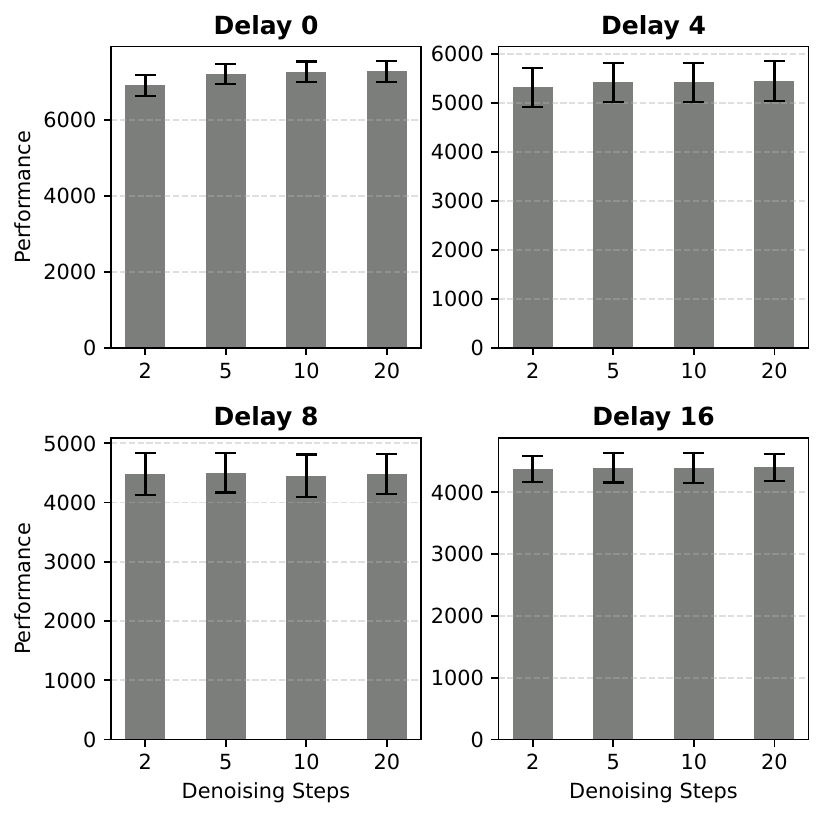}
    \caption{Examining SAID's performance for online RLwD (all-tasks average) with different denoising steps during inference. Data are Mean $\pm$ S.E.M.}
    \label{fig:denoise_steps}
    \vspace{-2mm}
\end{figure}

Secondly, we tested how diffusion denoising steps \cite{ho2020denoising} affects the effectiveness of our methods. It is well known that diffusion model makes more accurate and fine generation but suffers from linearly increased computational overhead when denoising step increases. Fig.~\ref{fig:denoise_steps} shows that SAID is also robust to the choice of denoising steps. The performance increases only slightly when more denoising steps are adopted.

\subsection{Inference Speed}

For decision making with delay, the time spend by the model itself is also, the above section has shown that our model performs similarly well even only with 2 denoising steps. In this case, we evaluated the inference latency of SAID using a single A-100 GPU. The latency at per environment step (single environment, delay=8) is approximately 0.015 seconds, which is much smaller than the robot's physical delay (Fig.~\ref{chap:online_results}, Bottom Right).

\section{Conclusion \& Discussion}
In this work, we introduced the State-Action Inpainting Diffuser (SAID). Rather than proposing a new model architecture, we propose a novel methodology that fundamentally reformulates RLwD as a trajectory inpainting task. By treating the unknown current states and future actions---masked by signal latency---as missing regions in a spatio-temporal sequence, SAID leverages the powerful generative capabilities of diffusion models to reconstruct optimal trajectories conditioned on delayed observations. The success of this approach is primarily driven by its ability to implicitly learn environmental dynamics without relying on an explicit, potentially biased world model. Furthermore, unlike autoregressive prediction methods that suffer from compounding errors, the diffusion-based inpainting mechanism effectively captures the uncertainty inherent in delayed control, allowing the agent to generate consistent and plausible plans even under significant latency.

SAID shares a notable limitation with other diffusion planning methods: the high computational cost during inference. The iterative denoising process required to generate high-fidelity trajectories introduces calculation latency that poses challenges for high-frequency real-time control systems. However, this computational bottleneck is not insurmountable. Recent advancements in diffusion acceleration, such as Habi~\citep{lu2025habitizing}, offer a promising solution by adapting the denoising process to reuse previous generation patterns. Incorporating such acceleration techniques could drastically reduce the number of required sampling steps, making SAID feasible for deployment on hardware with limited compute budgets without sacrificing the performance benefits of the inpainting formulation.

Future work will focus on extending the framework to handle highly stochastic or time-varying delays without retraining. Additionally, we aim to investigate the theoretical bounds of generative inpainting for control and validate the approach on physical robotic systems where delays are coupled with sensor noise. We believe the generative perspective introduced here establishes a new standard for robust continuous control, paving the way for delay-invariant autonomous systems.



\clearpage
\newpage

\section*{Impact Statement}
This research advances technical progress in machine learning by offering an effective decision-making framework under observation delay. It does not raise unique ethical issues beyond the usual considerations involved in the development and deployment of algorithms.

\bibliography{refenrence}

@inproceedings{revisitingstateaugmentation2021,
  title={Revisiting State Augmentation methods for Reinforcement Learning with Stochastic Delays},
  author={Nath, Somjit AND Baranwal, Mayank AND Khadilkar, Harshad},
  booktitle={ACM International Conference on Information and Knowledge Management},
  year={2021},
}

@inproceedings{learningbeliefrepresentation2021,
  title={Learning a Belief Representation for Delayed Reinforcement Learning},
  author={Liotet, Pierre AND Venneri, Erick AND Restelli, Marcello},
  booktitle={International Joint Conference on Neural Networks (IJCNN)},
  year={2021},
  doi={10.1109/IJCNN52387.2021.9534358},
}

@inproceedings{reinforcementlearningwith2021,
  title={Reinforcement learning with random delays},
  author={Bouteiller, Yann AND Ramstedt, Simon AND Beltrame, Giovanni AND Pal, Christopher J. AND Binas, Jonathan},
  booktitle={International Conference on Learning Representations},
  year={2021},
}

@inproceedings{delayedreinforcementlearning2022,
  title={Delayed reinforcement learning by imitation},
  author={Liotet, Paul AND Maran, Diego AND Bisi, Luciano AND Restelli, Marcello AND Pirotta, Matteo},
  booktitle={International Conference on Machine Learning},
  year={2022},
}

@inproceedings{addressingsignaldelay2024,
  title={Addressing Signal Delay in Deep Reinforcement Learning},
  author={Wang, Wei AND Han, Dongqi AND Luo, Xufang AND Li, Dongsheng},
  booktitle={International Conference on Learning Representations},
  year={2024},
}

@inproceedings{boostingreinforcementlearning2024,
  title={Boosting Reinforcement Learning with Strongly Delayed Feedback Through Auxiliary Short Delays},
  author={Wu, Qingyuan AND Zhan, Simon Sinong AND Wang, Yixuan AND Wang, Yuhui AND Lin, Chung-Wei AND Lv, Chen AND Zhu, Qi AND Schmidhuber, Jürgen AND Huang, Chao},
  booktitle={International Conference on Machine Learning},
  year={2024},
}

@inproceedings{variationaldelayedpolicy2024,
  title={Variational Delayed Policy Optimization},
  author={Wu, Qingyuan AND Zhan, Simon Sinong AND Wang, Yixuan AND Wang, Yuhui AND Lin, Chung-Wei AND Lv, Chen AND Zhu, Qi AND Schmidhuber, Jürgen AND Huang, Chao},
  booktitle={Advances in Neural Information Processing Systems},
  year={2024},
}

@article{reinforcementlearningfrom2024,
  title={Reinforcement Learning from Delayed Observations via World Models},
  author={Karamzade, Armin AND Kim, Kyungmin AND Kalsi, Montek AND Fox, Roy},
  journal={arXiv preprint arXiv:2403.12309},
  year={2024},
}

@inproceedings{directlyforecastingbelief2025,
  title={Directly Forecasting Belief for Reinforcement Learning with Delays},
  author={Wu, Qingyuan AND Wang, Yuhui AND Zhan, Simon Sinong AND Wang, Yixuan AND Lin, Chung-Wei AND Lv, Chen AND Zhu, Qi AND Schmidhuber, Jürgen AND Huang, Chao},
  booktitle={International Conference on Machine Learning},
  year={2025},
}

@inproceedings{ronneberger2015u,
  title={U-net: Convolutional networks for biomedical image segmentation},
  author={Ronneberger, Olaf and Fischer, Philipp and Brox, Thomas},
  booktitle={Medical Image Computing and Computer-Assisted Intervention--MICCAI 2015: 18th International Conference, Munich, Germany, October 5-9, 2015, Proceedings, Part III 18},
  pages={234--241},
  year={2015},
  organization={Springer}
}

@misc{fu2020d4rl,
    title={D4RL: Datasets for Deep Data-Driven Reinforcement Learning},
    author={Justin Fu and Aviral Kumar and Ofir Nachum and George Tucker and Sergey Levine},
    year={2020},
    eprint={2004.07219},
    archivePrefix={arXiv},
    primaryClass={cs.LG}
}

@article{agarwal2021blind,
  title={Blind decision making: Reinforcement learning with delayed observations},
  author={Agarwal, Mridul and Aggarwal, Vaneet},
  journal={Pattern Recognition Letters},
  volume={150},
  pages={176--182},
  year={2021},
  publisher={Elsevier}
}

@article{tarasov2023revisiting,
  title={Revisiting the minimalist approach to offline reinforcement learning},
  author={Tarasov, Denis and Kurenkov, Vladislav and Nikulin, Alexander and Kolesnikov, Sergey},
  journal={Advances in Neural Information Processing Systems},
  volume={36},
  pages={11592--11620},
  year={2023}
}

@article{zhan2025adapting,
  title={Adapting Offline Reinforcement Learning with Online Delays},
  author={Zhan, Simon Sinong and Wu, Qingyuan and Yang, Frank and Shi, Xiangyu and Huang, Chao and Zhu, Qi},
  journal={arXiv preprint arXiv:2506.00131},
  year={2025}
}

@inproceedings{
lu2025habitizing,
title={Habitizing Diffusion Planning for Efficient and Effective Decision Making},
author={Haofei Lu and Yifei Shen and Dongsheng Li and Junliang Xing and Dongqi Han},
booktitle={Forty-second International Conference on Machine Learning},
year={2025}
}

@article{towers2024gymnasium,
  title={Gymnasium: A standard interface for reinforcement learning environments},
  author={Towers, Mark and Kwiatkowski, Ariel and Terry, Jordan and Balis, John U and De Cola, Gianluca and Deleu, Tristan and Goul{\~a}o, Manuel and Kallinteris, Andreas and Krimmel, Markus and KG, Arjun and others},
  journal={arXiv preprint arXiv:2407.17032},
  year={2024}
}

@inproceedings{li2023hierarchical,
  title={Hierarchical diffusion for offline decision making},
  author={Li, Wenhao and Wang, Xiangfeng and Jin, Bo and Zha, Hongyuan},
  booktitle={International Conference on Machine Learning},
  pages={20035--20064},
  year={2023},
  organization={PMLR}
}

@inproceedings{chen2023offline,
  title={Offline Reinforcement Learning via High-Fidelity Generative Behavior Modeling},
  author={Chen, Huayu and Lu, Cheng and Ying, Chengyang and Su, Hang and Zhu, Jun},
  booktitle={The Eleventh International Conference on Learning Representations},
  year={2023}
}

@inproceedings{
han2020variational,
title={Variational Recurrent Models for Solving Partially Observable Control Tasks},
author={Dongqi Han and Kenji Doya and Jun Tani},
booktitle={International Conference on Learning Representations},
year={2020},
}

@inproceedings{
dong2024cleandiffuser,
title={CleanDiffuser: An Easy-to-use Modularized Library for Diffusion Models in Decision Making},
author={Zibin Dong and Yifu Yuan and Jianye HAO and Fei Ni and Yi Ma and Pengyi Li and YAN ZHENG},
booktitle={The Thirty-eight Conference on Neural Information Processing Systems Datasets and Benchmarks Track},
year={2024}
}

@article{chen2021delay,
  title={Delay-aware model-based reinforcement learning for continuous control},
  author={Chen, Baiming and Xu, Mengdi and Li, Liang and Zhao, Ding},
  journal={Neurocomputing},
  volume={450},
  pages={119--128},
  year={2021},
  publisher={Elsevier}
}

@article{boyan2000exact,
  title={Exact solutions to time-dependent MDPs},
  author={Boyan, Justin and Littman, Michael},
  journal={Advances in Neural Information Processing Systems},
  volume={13},
  year={2000}
}

@article{schulman2017proximal,
  title={Proximal policy optimization algorithms},
  author={Schulman, John and Wolski, Filip and Dhariwal, Prafulla and Radford, Alec and Klimov, Oleg},
  journal={arXiv preprint arXiv:1707.06347},
  year={2017}
}

@article{hansen2023idql,
  title={Idql: Implicit q-learning as an actor-critic method with diffusion policies},
  author={Hansen-Estruch, Philippe and Kostrikov, Ilya and Janner, Michael and Kuba, Jakub Grudzien and Levine, Sergey},
  journal={arXiv preprint arXiv:2304.10573},
  year={2023}
}

@inproceedings{ajay2022conditional,
  title={Is Conditional Generative Modeling all you need for Decision Making?},
  author={Ajay, Anurag and Du, Yilun and Gupta, Abhi and Tenenbaum, Joshua B and Jaakkola, Tommi S and Agrawal, Pulkit},
  booktitle={The Eleventh International Conference on Learning Representations},
  year={2022}
}

@article{liang2023adaptdiffuser,
  title={Adaptdiffuser: Diffusion models as adaptive self-evolving planners},
  author={Liang, Zhixuan and Mu, Yao and Ding, Mingyu and Ni, Fei and Tomizuka, Masayoshi and Luo, Ping},
  journal={ICML},
  year={2023}
}

@article{levine2020offline,
  title={Offline reinforcement learning: Tutorial, review, and perspectives on open problems},
  author={Levine, Sergey and Kumar, Aviral and Tucker, George and Fu, Justin},
  journal={arXiv preprint arXiv:2005.01643},
  year={2020}
}

@article{song2020denoising,
  title={Denoising diffusion implicit models},
  author={Song, Jiaming and Meng, Chenlin and Ermon, Stefano},
  journal={arXiv preprint arXiv:2010.02502},
  year={2020}
}

@inproceedings{fujimoto2019off,
  title={Off-policy deep reinforcement learning without exploration},
  author={Fujimoto, Scott and Meger, David and Precup, Doina},
  booktitle={International Conference on Machine Learning},
  pages={2052--2062},
  year={2019},
  organization={PMLR}
}

@inproceedings{
wang2023diffusion,
title={Diffusion Policies as an Expressive Policy Class for Offline Reinforcement Learning},
author={Zhendong Wang and Jonathan J Hunt and Mingyuan Zhou},
booktitle={The Eleventh International Conference on Learning Representations },
year={2023}
}

@inproceedings{janner2022planning,
  title={Planning with Diffusion for Flexible Behavior Synthesis},
  author={Janner, Michael and Du, Yilun and Tenenbaum, Joshua and Levine, Sergey},
  booktitle={International Conference on Machine Learning},
  pages={9902--9915},
  year={2022},
  organization={PMLR}
}

@article{ho2020denoising,
  title={Denoising diffusion probabilistic models},
  author={Ho, Jonathan and Jain, Ajay and Abbeel, Pieter},
  journal={Advances in neural information processing systems},
  volume={33},
  pages={6840--6851},
  year={2020}
}

@article{vinyals2019grandmaster,
  title={Grandmaster level in StarCraft II using multi-agent reinforcement learning},
  author={Vinyals, Oriol and Babuschkin, Igor and Czarnecki, Wojciech M and Mathieu, Micha{\"e}l and Dudzik, Andrew and Chung, Junyoung and Choi, David H and Powell, Richard and Ewalds, Timo and Georgiev, Petko and others},
  journal={Nature},
  volume={575},
  number={7782},
  pages={350--354},
  year={2019},
  publisher={Nature Publishing Group}
}

@inproceedings{schmidhuber1991reinforcement,
  title={Reinforcement learning in {M}arkovian and non-{M}arkovian environments},
  author={Schmidhuber, J{\"u}rgen},
  booktitle={Advances in Neural Information Processing Systems},
  pages={500--506},
  year={1991}
}

@inproceedings{peebles2023scalable,
  title={Scalable diffusion models with transformers},
  author={Peebles, William and Xie, Saining},
  booktitle={Proceedings of the IEEE/CVF International Conference on Computer Vision},
  pages={4195--4205},
  year={2023}
}

@article{kostrikov2021offline,
  title={Offline reinforcement learning with implicit q-learning},
  author={Kostrikov, Ilya and Nair, Ashvin and Levine, Sergey},
  journal={arXiv preprint arXiv:2110.06169},
  year={2021}
}

@article{kumar2020conservative,
  title={Conservative q-learning for offline reinforcement learning},
  author={Kumar, Aviral and Zhou, Aurick and Tucker, George and Levine, Sergey},
  journal={Advances in Neural Information Processing Systems},
  volume={33},
  pages={1179--1191},
  year={2020}
}

@inproceedings{lu2025what,
  title={What Makes a Good Diffusion Planner for Decision Making?},
  author={Haofei Lu and Dongqi Han and Yifei Shen and Dongsheng Li},
  booktitle={The Thirteenth International Conference on Learning Representations},
  year={2025},
}

@article{ren2024diffusion,
  title={Diffusion policy policy optimization},
  author={Ren, Allen Z and Lidard, Justin and Ankile, Lars L and Simeonov, Anthony and Agrawal, Pulkit and Majumdar, Anirudha and Burchfiel, Benjamin and Dai, Hongkai and Simchowitz, Max},
  journal={arXiv preprint arXiv:2409.00588},
  year={2024}
}

@article{psenka2023learning,
  title={Learning a diffusion model policy from rewards via {Q}-score matching},
  author={Psenka, Michael and Escontrela, Alejandro and Abbeel, Pieter and Ma, Yi},
  journal={arXiv preprint arXiv:2312.11752},
  year={2023}
}

@article{yang2023policy,
  title={Policy representation via diffusion probability model for reinforcement learning},
  author={Yang, Long and Huang, Zhixiong and Lei, Fenghao and Zhong, Yucun and Yang, Yiming and Fang, Cong and Wen, Shiting and Zhou, Binbin and Lin, Zhouchen},
  journal={arXiv preprint arXiv:2305.13122},
  year={2023}
}

@article{wang2024diffusion,
  title={Diffusion Actor-Critic with Entropy Regulator},
  author={Wang, Yinuo and Wang, Likun and Jiang, Yuxuan and Zou, Wenjun and Liu, Tong and Song, Xujie and Wang, Wenxuan and Xiao, Liming and Wu, Jiang and Duan, Jingliang and others},
  journal={arXiv preprint arXiv:2405.15177},
  year={2024}
}

@article{ze20243d,
  title={3{D} diffusion policy},
  author={Ze, Yanjie and Zhang, Gu and Zhang, Kangning and Hu, Chenyuan and Wang, Muhan and Xu, Huazhe},
  journal={arXiv preprint arXiv:2403.03954},
  year={2024}
}

@article{chi2023diffusion,
  title={Diffusion policy: Visuomotor policy learning via action diffusion},
  author={Chi, Cheng and Xu, Zhenjia and Feng, Siyuan and Cousineau, Eric and Du, Yilun and Burchfiel, Benjamin and Tedrake, Russ and Song, Shuran},
  journal={The International Journal of Robotics Research},
  pages={02783649241273668},
  year={2023},
  publisher={SAGE Publications Sage UK: London, England}
}

@article{paszke2019pytorch,
  title={Pytorch: An imperative style, high-performance deep learning library},
  author={Paszke, Adam and Gross, Sam and Massa, Francisco and Lerer, Adam and Bradbury, James and Chanan, Gregory and Killeen, Trevor and Lin, Zeming and Gimelshein, Natalia and Antiga, Luca and others},
  journal={Advances in neural information processing systems},
  volume={32},
  year={2019}
}

@article{yang2022rorl,
  title={Rorl: Robust offline reinforcement learning via conservative smoothing},
  author={Yang, Rui and Bai, Chenjia and Ma, Xiaoteng and Wang, Zhaoran and Zhang, Chongjie and Han, Lei},
  journal={Advances in neural information processing systems},
  volume={35},
  pages={23851--23866},
  year={2022}
}

@article{an2021uncertainty,
  title={Uncertainty-based offline reinforcement learning with diversified q-ensemble},
  author={An, Gaon and Moon, Seungyong and Kim, Jang-Hyun and Song, Hyun Oh},
  journal={Advances in neural information processing systems},
  volume={34},
  pages={7436--7447},
  year={2021}
}

@article{abadia2021cerebellar,
  title={A cerebellar-based solution to the nondeterministic time delay problem in robotic control},
  author={Abad{\'\i}a, Ignacio and Naveros, Francisco and Ros, Eduardo and Carrillo, Richard R and Luque, Niceto R},
  journal={Science Robotics},
  volume={6},
  number={58},
  pages={eabf2756},
  year={2021},
  publisher={American Association for the Advancement of Science}
}

@article{bastian2006learning,
  title={Learning to predict the future: the cerebellum adapts feedforward movement control},
  author={Bastian, Amy J},
  journal={Current opinion in neurobiology},
  volume={16},
  number={6},
  pages={645--649},
  year={2006},
  publisher={Elsevier}
}

@article{haarnoja2019soft,
  title={Soft actor-critic algorithms and applications},
  author={Haarnoja, Tuomas and Zhou, Aurick and Hartikainen, Kristian and Tucker, George and Ha, Sehoon and Tan, Jie and Kumar, Vikash and Zhu, Henry and Gupta, Abhishek and Abbeel, Pieter and others},
  journal={arXiv preprint arXiv:1812.05905},
  year={2018}
}

@article{bellman1957markovian,
  title={A {M}arkovian decision process},
  author={Bellman, Richard},
  journal={Journal of Mathematics and Mechanics},
  pages={679--684},
  year={1957},
  publisher={JSTOR}
}

@inproceedings{haarnoja2018soft,
  title={Soft Actor-Critic: Off-Policy Maximum Entropy Deep Reinforcement Learning with a Stochastic Actor},
  author={Haarnoja, Tuomas and Zhou, Aurick and Abbeel, Pieter and Levine, Sergey},
  booktitle={International Conference on Machine Learning},
  pages={1856--1865},
  year={2018}
}

@article{degrave2022magnetic,
  title={Magnetic control of {T}okamak plasmas through deep reinforcement learning},
  author={Degrave, Jonas and Felici, Federico and Buchli, Jonas and Neunert, Michael and Tracey, Brendan and Carpanese, Francesco and Ewalds, Timo and Hafner, Roland and Abdolmaleki, Abbas and de Las Casas, Diego and others},
  journal={Nature},
  volume={602},
  number={7897},
  pages={414--419},
  year={2022},
  publisher={Nature Publishing Group}
}

@Book{sutton1998reinforcement,
  title     = {Reinforcement learning: An introduction},
  publisher = {MIT press Cambridge},
  year      = {1998},
  author    = {Sutton, Richard S and Barto, Andrew G},
  volume    = {1},
}

@inproceedings{meng2004remote,
  title={Remote surgery case: robot-assisted teleneurosurgery},
  author={Meng, Cai and Wang, Tianmiao and Chou, Wusheng and Luan, Sheng and Zhang, Yuru and Tian, Zengmin},
  booktitle={IEEE International Conference on Robotics and Automation, 2004. Proceedings. ICRA'04. 2004},
  volume={1},
  pages={819--823},
  year={2004},
  organization={IEEE}
}

@inproceedings{fang2021universal,
  title={Universal trading for order execution with oracle policy distillation},
  author={Fang, Yuchen and Ren, Kan and Liu, Weiqing and Zhou, Dong and Zhang, Weinan and Bian, Jiang and Yu, Yong and Liu, Tie-Yan},
  booktitle={Proceedings of the AAAI Conference on Artificial Intelligence},
  volume={35},
  number={1},
  pages={107--115},
  year={2021}
}

@article{gerwig2005timing,
  title={Timing of conditioned eyeblink responses is impaired in cerebellar patients},
  author={Gerwig, Marcus and Hajjar, Karim and Dimitrova, Albena and Maschke, Matthias and Kolb, Florian P and Frings, Markus and Thilmann, Alfred F and Forsting, Michael and Diener, Hans Christoph and Timmann, Dagmar},
  journal={Journal of Neuroscience},
  volume={25},
  number={15},
  pages={3919--3931},
  year={2005},
  publisher={Soc Neuroscience}
}

@article{jafaripournimchahi2022stability,
  title={Stability analysis of delayed-feedback control effect in the continuum traffic flow of autonomous vehicles without V2I communication},
  author={Jafaripournimchahi, Ammar and Cai, Yingfeng and Wang, Hai and Sun, Lu and Yang, Biao},
  journal={Physica A: Statistical Mechanics and its Applications},
  volume={605},
  pages={127975},
  year={2022},
  publisher={Elsevier}
}

@inproceedings{hafner2019dream,
  title={Dream to Control: Learning Behaviors by Latent Imagination},
  author={Hafner, Danijar and Lillicrap, Timothy and Ba, Jimmy and Norouzi, Mohammad},
  booktitle={International Conference on Learning Representations},
  year={2019}
}

@article{mnih2015human,
  title={Human-level control through deep reinforcement learning},
  author={Mnih, Volodymyr and Kavukcuoglu, Koray and Silver, David and Rusu, Andrei A and Veness, Joel and Bellemare, Marc G and Graves, Alex and Riedmiller, Martin and Fidjeland, Andreas K and Ostrovski, Georg and others},
  journal={Nature},
  volume={518},
  number={7540},
  pages={529},
  year={2015},
  publisher={Nature Publishing Group}
}

@article{brown2020language,
  title={Language models are few-shot learners},
  author={Brown, Tom and Mann, Benjamin and Ryder, Nick and Subbiah, Melanie and Kaplan, Jared D and Dhariwal, Prafulla and Neelakantan, Arvind and Shyam, Pranav and Sastry, Girish and Askell, Amanda and others},
  journal={Advances in neural information processing systems},
  volume={33},
  pages={1877--1901},
  year={2020}
}
\bibliographystyle{icml2025}

\clearpage
\appendix
\onecolumn

\section{Algorithm Pseudo-Code}

\begin{algorithm*}[h]
   \caption{SAID for Online RLwD}
   \label{alg:online_rl}
\begin{algorithmic}[1]
    \State {\bfseries Input:} Task environment $\mathcal{E}$ with observation delay $\Delta_t$
   \State {\bfseries Input:} Randomly initialized diffusion model $\bm\theta$, planning horizon $H$
   \State {\bfseries Input:} Any online RLwD algorithm $\mathcal{A}$ (we adopt \citet{addressingsignaldelay2024} in this study) and its policy model $\pi$ and state value function $v$; replay buffer $\mathcal{B}$
   
   \State \Comment{\textcolor{gray}{\textit{Pre-learning and collect data}}}
   
   \For{episode $m = 1$ to $M$}
       \State Reset environment, initial observation $\tilde{s}_0$.
       \State $t \leftarrow 0$
       
       \While{not terminal or truncated}
           \State Sample action $a_t \sim \pi(a|s_t,a_{t-\Delta_t:t-1})$ 
           
           \State Execute $a_t$ in $\mathcal{E}$
           
           \State Receive reward $r_t$ and next observation $\tilde{s}_{t+1}$
           
           \State Store transition $(\tilde{s}_t, a_t, r_t, \tilde{s}_{t+1})$ in $\mathcal{B}$

           \State Update policy $\pi$ and value function $v$ using $\mathcal{A}$
           
           \State $t \leftarrow t + 1$, $s_t \leftarrow s_{t+1}$
       \EndWhile
   \EndFor
   
   \State \Comment{\textcolor{gray}{\textit{Train diffusion planner of SAID}}}
   
   \For{gradient step $n = 1$ to $N$}
    \State Randomly sample a batch of state-action sequences $(\tilde{s}_{\tau:\tau + H - 1}, a_{\tau-\Delta_t:\tau-\Delta + H - 1})$ from $\mathcal{B}$
    \State Train $\bm\theta$ conditioned on $(\tilde{s}_\tau, a_{\tau-\Delta_t:{\tau - 1}})$ (Fig.~\ref{fig:method}) (Here we adopted DV as base model \citep{lu2025what})
   \EndFor

   \State \Comment{\textcolor{gray}{\textit{Inference using SAID}}}
   
   \For{evaluation episode $m' = 1$ to $M'$}
       \State Reset environment, initial observation $\tilde{s}_0$.
       \State $t \leftarrow 0$
       \State Initialize action buffer $\mathcal{D}$
       \While{not terminal or truncated}

           \State Use $\theta$ to generate multiple candidates of $\tilde{s}_{t+1:t+H+1}, a_{t:t+H-1}$, conditioned on $(\tilde{s}_t, a_{t -\Delta_t:{t - 1}})$ (Fig.~\ref{fig:method},\ref{fig:init_steps})

           \State Compute the mean value of $\tilde{s}_{t+1:t+H+1}$ using trained value function $v$ and select the best trajectory from all the candidates
           
           \State Choose $a_t$ from the best trajectory in $\mathcal{E}$
           \State Append $a_t$ to $\mathcal{D}$
           \State Receive reward $r_t$ an next observation $\title{s}_{t+1}$
           
           \State $t \leftarrow t + 1$,  \State $s_t \leftarrow s_{t+1}$
       \EndWhile
   \EndFor

\end{algorithmic}
\end{algorithm*}

\begin{algorithm*}[h]
   \caption{SAID for offline RLwD}
   \label{alg:offline_rl}
\begin{algorithmic}[1]
    \State {\bfseries Input:} Task environment $\mathcal{E}$, offline dataset $(s_t, a_t, r_t, s_{t+1})$ , and observation delay $\Delta_t$.
   \State {\bfseries Input:} Randomly initialized diffusion model $\bm\theta$, planning horizon $H$, value function $v$ and corresponding learning algorithm (here we adopt implicit Q-learning \citep{kostrikov2021offline})
   
   \State \Comment{\textcolor{gray}{\textit{Pre-learning and collect data}}}
    \State Convert offline dataset to $(\tilde{s}_t, a_t, r_t, \tilde{s}_{t+1})$ by shifting the states by $\Delta_t$ steps.
   
   \State \Comment{\textcolor{gray}{\textit{Train diffusion planner of SAID and value function}}}
   
   \For{gradient step $n = 1$ to $N$}
    \State Randomly sample a batch of state-action sequences $(\tilde{s}_{\tau:\tau + H - 1}, a_{\tau-\Delta_t:\tau-\Delta + H - 1})$ from $\mathcal{B}$
    \State Train $\bm{\theta}$ conditioned on $(\tilde{s}_\tau, a_{\tau-\Delta_t:{\tau - 1}})$ (Fig.~\ref{fig:method}) 
    \State Train value function $v$
   \EndFor

   \State \Comment{\textcolor{gray}{\textit{Inference using SAID}}}
   
   \For{evaluation episode $m' = 1$ to $M'$}
       \State Reset environment, initial observation $\tilde{s}_0$.
       \State $t \leftarrow 0$
       \State Initialize action buffer $\mathcal{D}$
       \While{not terminal or truncated}

           \State Use $\theta$ to generate multiple candidates of $\tilde{s}_{t+1:t+H+1}, a_{t:t+H-1}$, conditioned on $(\tilde{s}_t, a_{t -\Delta_t:{t - 1}})$ (Fig.~\ref{fig:method},\ref{fig:init_steps})

           \State Compute the mean value of $\tilde{s}_{t+1:t+H+1}$ using trained value function $v$ and select the best trajectory from all the candidates
           
           \State Choose $a_t$ from the best trajectory
           \State Execute $a_t$ in $\mathcal{E}$; append $a_t$ to $\mathcal{D}$
           \State Receive reward $r_t$ an next observation $\title{s}_{t+1}$
           
           \State $t \leftarrow t + 1$,  \State $s_t \leftarrow s_{t+1}$
       \EndWhile
   \EndFor

\end{algorithmic}
\end{algorithm*}

\clearpage
\newpage

\section{Implementation Details}

\label{appendix:implementation}

Our implementation is built upon the PyTorch framework \cite{paszke2019pytorch}, utilizing the \texttt{cleandiffuser} library \cite{dong2024cleandiffuser} for the diffusion model backbone and sampling utilities. 

\subsection{Environment and Signal Delay}
To simulate observation delays in both online and offline settings, we implemented a custom environment wrapper, \texttt{DelayedEnvWrapper}. This wrapper maintains a First-In-First-Out (FIFO) buffer with a capacity equal to the delay steps $\Delta t$.
\begin{itemize}
    \item \textbf{Initialization:} Upon environment reset, the buffer is initialized by repeating the initial observation $s_0$ for $\Delta t$ times.
    \item \textbf{Step Function:} At each timestep $t$, the wrapper returns the delayed observation $s_{t-\Delta t}$ from the front of the buffer and appends the newly received true observation $s_{t+1}$ to the back.
    \item \textbf{Reward and Termination:} Consistent with the problem formulation, rewards $r_t$ and termination signals are provided without delay.
\end{itemize}

\subsection{Dataset and Preprocessing}
For offline reinforcement learning, we utilize the D4RL benchmark datasets. For online experiments, we use Gymnasium environments (\texttt{HalfCheetah-v5}, \texttt{Ant-v5}, \texttt{Hopper-v5}, \texttt{Walker2d-v5}) \cite{towers2024gymnasium}.
\begin{itemize}
    \item \textbf{Sequence Construction:} We utilize a sequence dataset loader. The planning horizon is dynamically adjusted to $H + \Delta t$, ensuring the model captures both the delay window and the future planning horizon.
    \item \textbf{Normalization:} States and actions are normalized to zero mean and unit variance. Actions are clipped to $[-0.999, 0.999]$ and transformed using the inverse hyperbolic tangent ($\text{arctanh}$) to map them to an unbounded Gaussian distribution suitable for diffusion training.
\end{itemize}

\subsection{Model Architecture}
The core of the State-Action Inpainting Diffuser (SAID) is a conditional diffusion model designed to recover the sequence of true states and actions.
\begin{itemize}
    \item \textbf{Backbone:} We employ a 1D Diffusion Transformer (\texttt{DiT1d}) as the noise prediction network $\epsilon_\theta$. The network processes the flattened state-action trajectories. We utilize Fourier feature encodings for timestep embeddings. The architecture depth are configurable hyperparameters (see Appendix~\ref{appendix:hps}), and the network width is 256.
    \item \textbf{Inpainting Mask:} We utilize a fixed masking strategy (`fix\_mask`) to enforce conditions. The current observation (delayed state) $\tilde{s}_t$ (at relative index 0) and the history of executed actions $a_{t-\Delta t : t-1}$ (indices $0$ to $\Delta t$) are masked as known values. The diffusion model is tasked with inpainting the unknown future states and actions.
\end{itemize}

\subsection{Training}
The diffusion planner is trained to approximate the data distribution of valid trajectories.
\begin{itemize}
    \item \textbf{Optimization:} We use the Adam optimizer with a Cosine Annealing learning rate scheduler.
    \item \textbf{Loss Function:} We support a weighted regression loss to prioritize high-value trajectories. The loss is computed as:
    \begin{equation}
        \mathcal{L} = \mathbb{E} \left[ w(\tau) \cdot || \epsilon - \epsilon_\theta(x_t, t) ||^2 \right]
    \end{equation}
    where the weight $w(\tau) = \exp(\beta \cdot (V(\tau) - 1))$ is derived from the value estimates of the trajectory, controlled by a temperature parameter $\beta$ (weight\_factor).
    \item \textbf{Value Function:} For offline RL, we concurrently train an Implicit Q-Learning (IQL) \cite{kostrikov2021offline} value function ($V_\psi$) to provide the necessary value guidance and trajectory evaluation using MCSS \citep{lu2025what}.
\end{itemize}

\subsection{Inference and Control}
During evaluation, the model operates in a receding horizon control loop:
\begin{itemize}
    \item \textbf{Sampling:} At each step, we generate $N$ candidate trajectories in parallel (where $N = \texttt{num\_envs} \times \texttt{planner\_num\_candidates}$). The sampling process begins with Gaussian noise and is denoised using a DDIM scheduler.
    \item \textbf{Conditioning:} The reverse diffusion process is conditioned on the delayed observation history and previously executed actions using the inpainting mask.
    \item \textbf{Selection Strategy (MCSS):} We employ Monte-Carlo Sampling with Selection. The generated candidate trajectories are evaluated using the learned Value Network $V_\psi$. The trajectory with the highest mean predicted value is selected as the optimal plan.
    \item \textbf{Execution:} The action corresponding to the current time step (relative index $\Delta t$ in the generated sequence) is extracted, denormalized (applying $\tanh$), and executed in the environment.
\end{itemize}

For other potentially not mentioned configurations, we simply follow the SOTA diffusion planner DV \cite{lu2025what}.

\subsection{Hyper-parameters}
\label{appendix:hps}

Most hyper-parameters are task-independent, which are specified in Table~\ref{table:shared_hp}.

\begin{table}[h]
\centering
\resizebox{1.0\textwidth}{!}{
\begin{tblr}{
  hline{1,20} = {-}{0.08em},
  hline{2} = {-}{},
}
Area                         & Hyperparameter                      & Value           & What it controls~                             \\
Environment / reward shaping & terminal\_penalty                   & -100            & Penalty on terminal condition.                \\
Environment / reward shaping & full\_traj\_bonus                   & 0               & Bonus for completing full trajectory.         \\
Planner diffusion            & planner\_solver                     & ddim            & Sampling/solver for planner diffusion.        \\
Planner architecture         & planner\_emb\_dim                   & 128             & Planner embedding dimension.                  \\
Planner architecture         & planner\_d\_model                   & 256             & Transformer model width (d\_model).           \\
Planner sampling             & planner\_sampling\_steps            & 20              & Steps used when sampling planner diffusion.   \\
Planner objective            & planner\_predict\_noise             & TRUE            & Train planner to predict noise.               \\
Planner EMA                  & planner\_ema\_rate                  & 0.9999          & EMA decay rate for planner parameters.        \\
Regression weighting         & weight\_factor~                     & 2               & Strength for weight regression                \\
Critic learning              & discount                            & 0.99            & Discount factor                               \\
Critic learning              & critic\_learning\_rate              & 0.0003          & Critic function learning rate                 \\
Critic learning              & iql\_tau                            & 0.7             & Hyper-parameter $\tau$ in IQL (for offline RL)\\
Training budget              & planner\_diffusion\_gradient\_steps & 1000000         & Update steps for planner diffusion.           \\
Training batch               & batch\_size                         & 256             & Batch size for training.                      \\
Planner inference            & planner\_num\_candidates            & 50              & Number of candidate plans sampled/considered. \\
Planner inference            & planner\_use\_ema                   & TRUE            & Use EMA weights for planner at inference.     \\
Planning                     & planner\_horizon                    & 8 + Delay Steps & Plan sequence length                          \\
Planning                     & stride                              & 1               & Plan sequence stride~                         
\end{tblr}
}
\caption{Task-independent hyperparameters of SAID.}
\label{table:shared_hp}
\end{table}

We did grid search for two crucial hyper-parameters (planner network depth and inference temperature). The importance of these two hyper-parameters have also been discussed in previous study by \citet{lu2025what}.The search range was $[0, 0.3, 0.6, 0.8, 1]$ for temperature and $[2, 4, 8]$ for planner depth. The tunned hyper-parameters are reported in Table~\ref{table:hps}.

\begin{table}[h]
\centering
\small
\begin{tblr}{
  cells = {c},
  cell{2}{1} = {c=5}{},
  cell{7}{1} = {c=5}{},
  cell{11}{1} = {c=5}{},
  cell{15}{1} = {c=5}{},
  hline{1,19} = {-}{0.08em},
  hline{2,7,11,15} = {-}{},
}
Task
  Name          & Delay 0 (pd / tmp) & Delay 4 (pd / tmp) & Delay 8 (pd / tmp) & Delay 16 (pd / tmp) \\
Online RL Tasks      &                    &                    &                    &                     \\
Ant-v5               & 8 / 0.6            & 8 / 0.6            & 8 / 0.6            & 8 / 0.8             \\
HalfCheetah-v5       & 8 / 0.6            & 8 / 0.3            & 8 / 0.3            & 8 / 0.1             \\
Hopper-v5            & 8 / 0.6            & 4 / 0.3            & 2 / 0.1            & 8 / 0.6             \\
Walker2d-v5          & 8 / 0.8            & 8 / 0.6            & 4 / 0.8            & 8 / 0.6             \\
HalfCheetah
  (D4RL) &                    &                    &                    &                     \\
medium-expert-v2     & 8 / 0.8            & 8 / 0.3            & 8 / 0.1            & 4 / 0.1             \\
medium-replay-v2     & 4 / 1.0            & 8 / 0.6            & 8 / 0.1            & 8 / 0.3             \\
medium-v2            & 8 / 0.8            & 8 / 0.3            & 8 / 0.1            & 8 / 0.1             \\
Walker2d
  (D4RL)    &                    &                    &                    &                     \\
medium-expert-v2     & 8 / 1.0            & 8 / 0.8            & 8 / 0.8            & 8 / 0.3             \\
medium-replay-v2     & 2 / 0.8            & 4 / 0.8            & 4 / 0.6            & 8 / 0.8             \\
medium-v2            & 8 / 1.0            & 8 / 1.0            & 4 / 1.0            & 8 / 0.6             \\
Hopper
  (D4RL)      &                    &                    &                    &                     \\
medium-expert-v2     & 8 / 0.3            & 8 / 0.3            & 4 / 0.1            & 8 / 0.1             \\
medium-replay-v2     & 2 / 0.6            & 8 / 1.0            & 8 / 0.8            & 4 / 0.8             \\
medium-v2            & 8 / 1.0            & 4 / 1.0            & 4 / 1.0            & 8 / 1.0             
\end{tblr}
\caption{Best hyperparameters of SAID for each task after grid search (pd: depth of the planner DiT, tmp: temperature of the planner during inference).}
\label{table:hps}
\end{table}

\section{Extensive Experimental Results}

\begin{figure}
    \centering
    \includegraphics[width=0.8\linewidth]{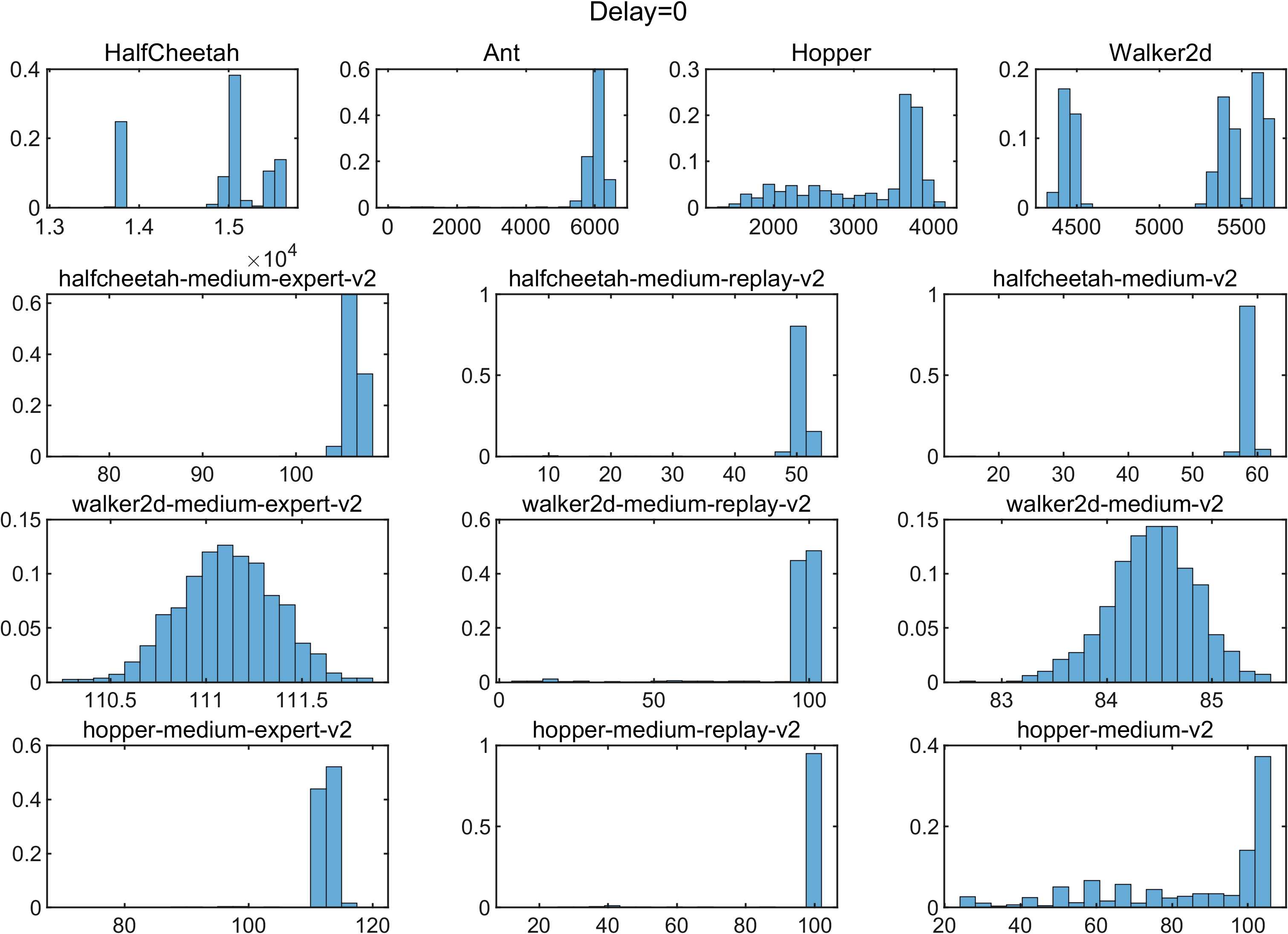}
    \caption{Episodic returns of SAID for each task (delay=0). The histogram shows the performance distribution of 200 episodes of one trained diffusion planning model.}
    \label{fig:all_d0}
\end{figure}

\begin{figure}
    \centering
    \includegraphics[width=0.8\linewidth]{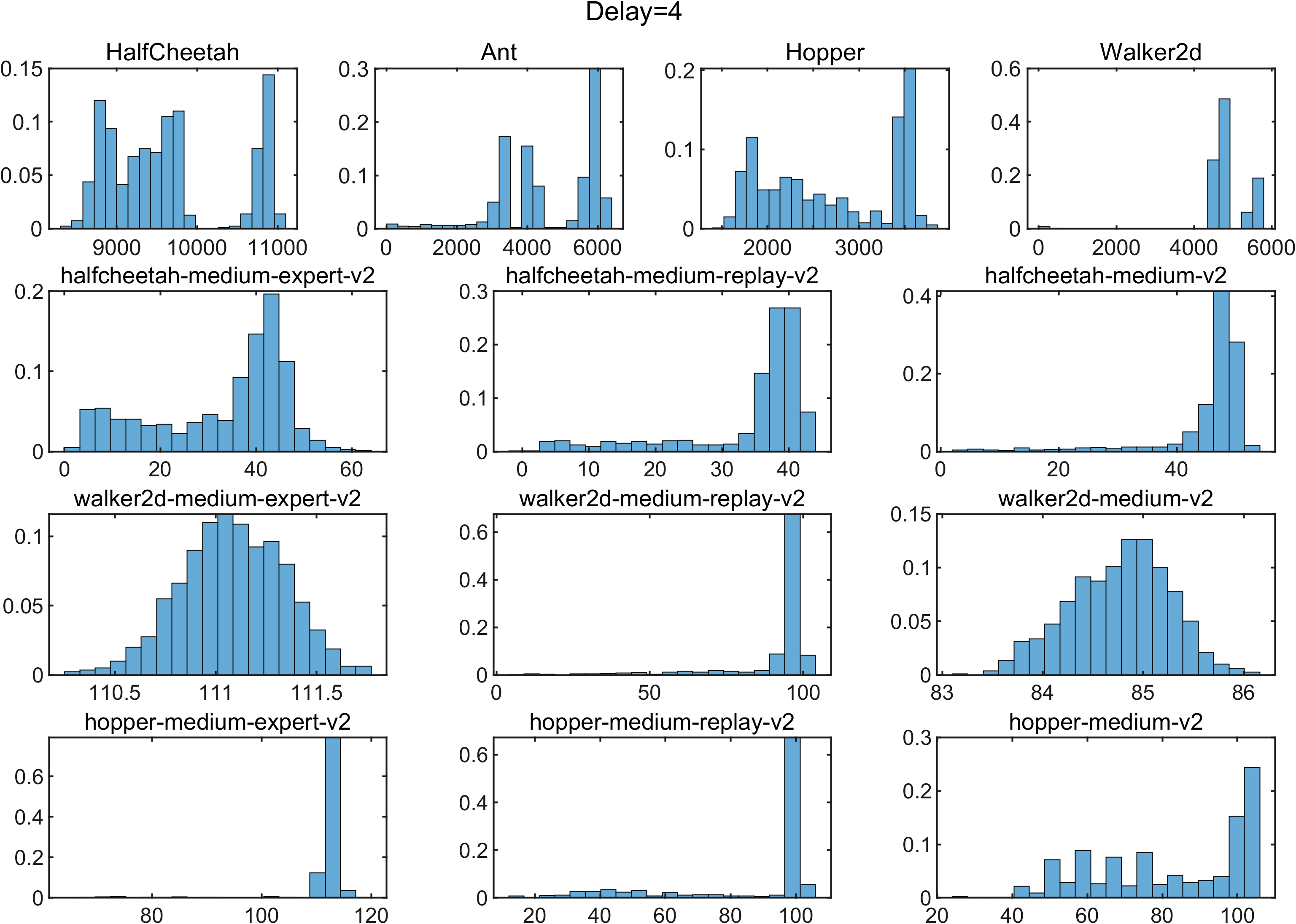}
    \caption{Episodic returns of SAID for each task (delay=4). The histogram shows the performance distribution of 200 episodes of one trained diffusion planning model.}
    \label{fig:all_d4}
\end{figure}

\begin{figure}
    \centering
    \includegraphics[width=0.8\linewidth]{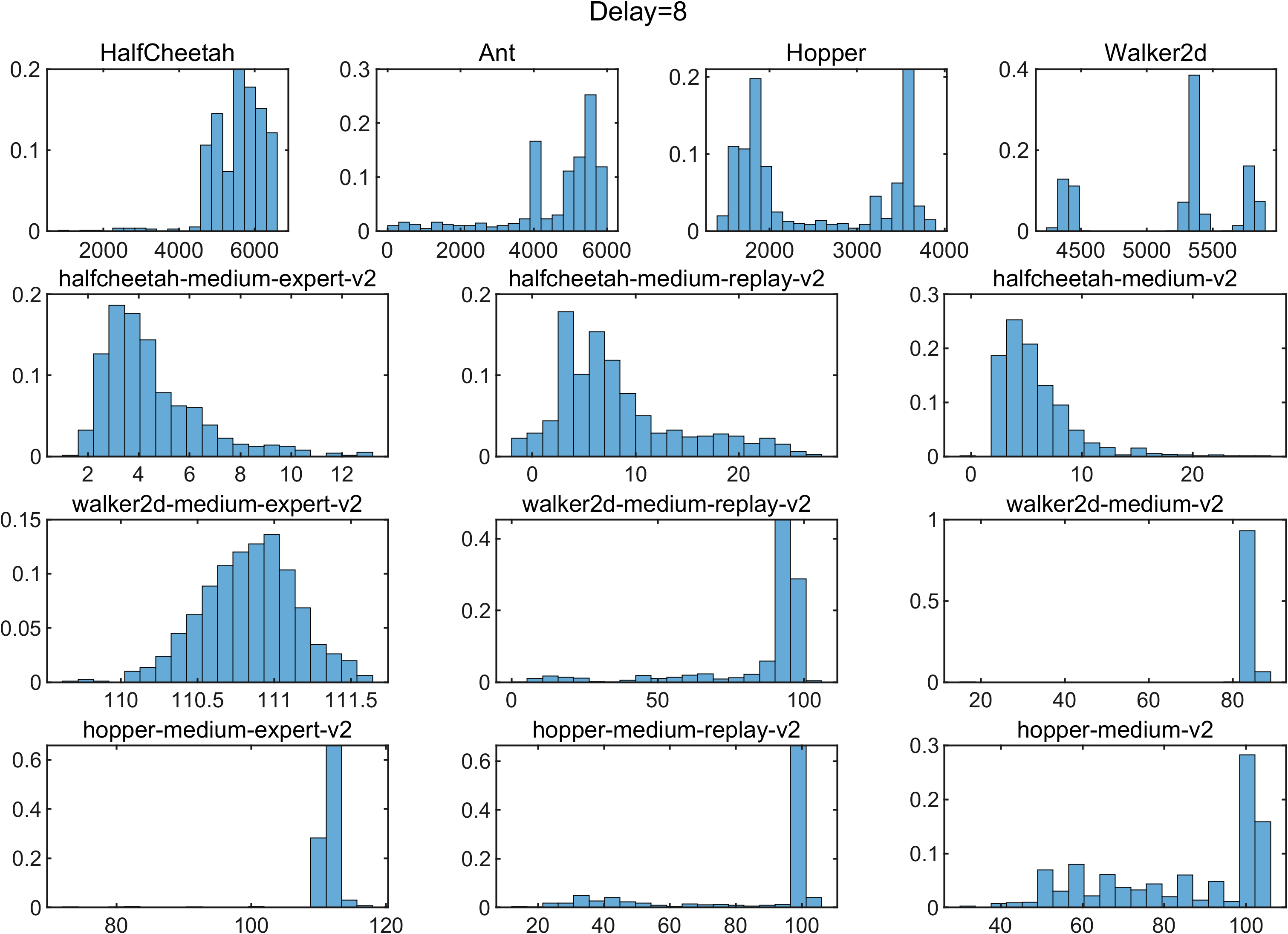}
    \caption{Episodic returns of SAID for each task (delay=8). The histogram shows the performance distribution of 200 episodes of one trained diffusion planning model.}
    \label{fig:all_d8}
\end{figure}

\begin{figure}
    \centering
    \includegraphics[width=0.8\linewidth]{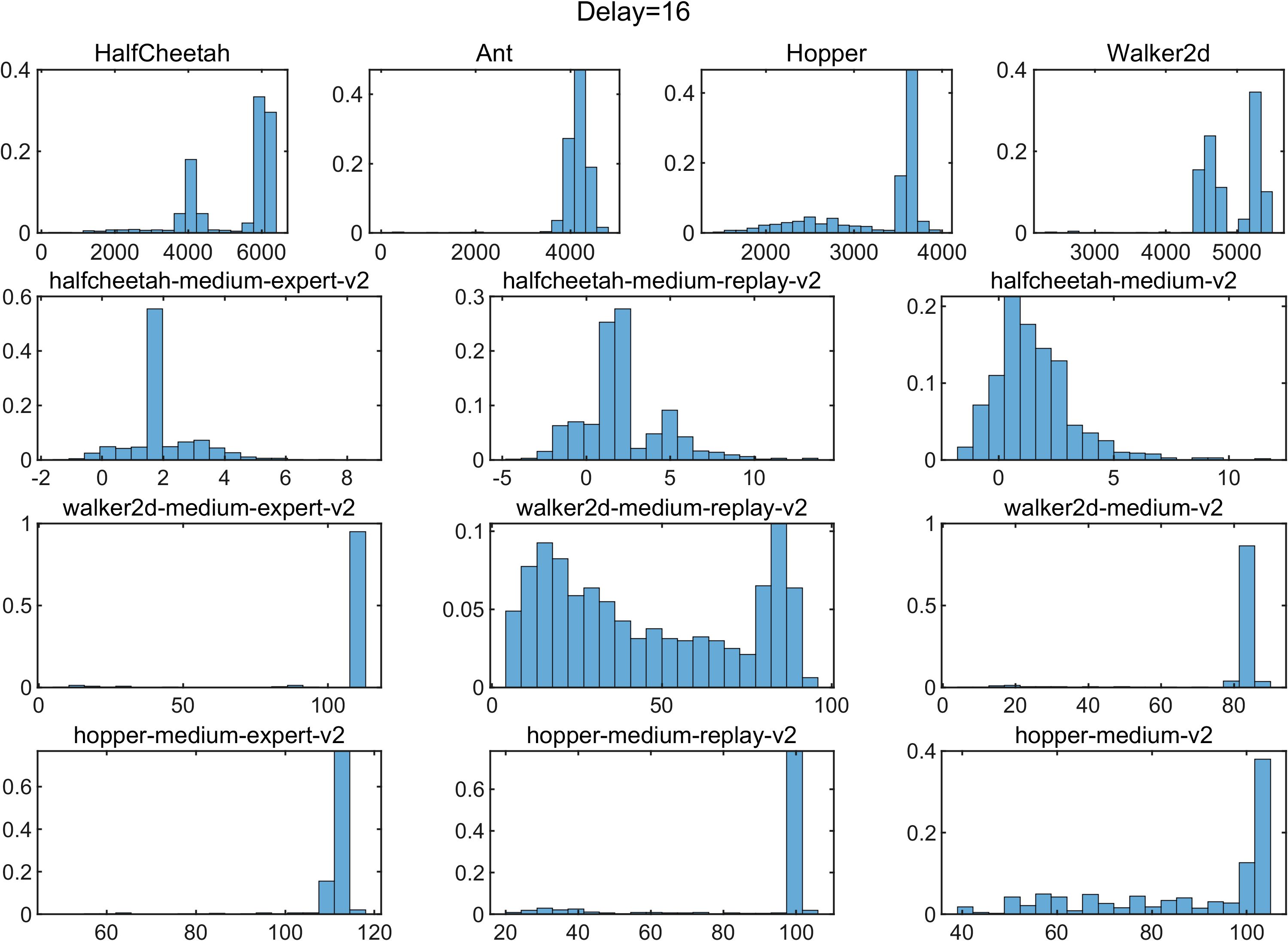}
    \caption{Episodic returns of SAID for each task (delay=16). The histogram shows the performance distribution of 200 episodes of one trained diffusion planning model.}
    \label{fig:all_d16}
\end{figure}

We also provide the SAID’s offline RL performance with error bar (Table.~\ref{table:offline_results_supp}, the detailed performance when denoising steps changes (Table.~\ref{table:denoising_steps}), as well as the distributions of episodic returns of SAID for each task (Fig.~\ref{fig:all_d0},\ref{fig:all_d4},\ref{fig:all_d8},\ref{fig:all_d16}).

\begin{figure}
    \centering
    \includegraphics[width=0.7\linewidth]{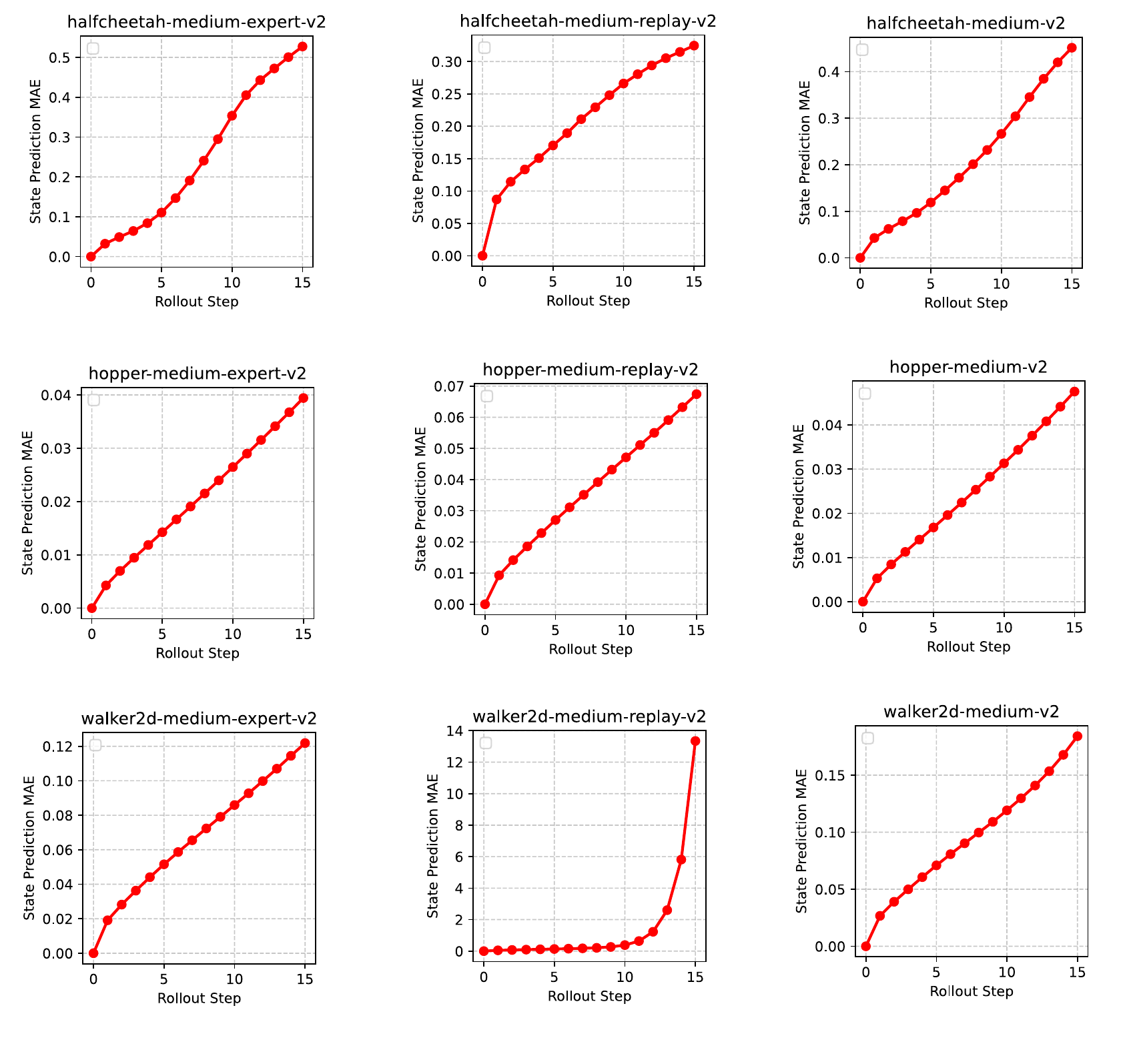}
    \caption{Full D4RL locomotion task results for Fig.~\ref{fig:motivation}.}
    \label{fig:compound}
\end{figure}

\begin{table}
\centering
\begin{tblr}{
  cells = {c},
  hline{1,11} = {-}{0.08em},
  hline{2} = {-}{},
}
Task                         & Delay 0     & Delay 4     & Delay 8     & Delay 16    \\
halfcheetah-medium-expert-v2 & 106.2 ± 0.1 & 32.8 ± 0.8  & 4.5 ± 0.1   & 1.9 ± 0.5   \\
halfcheetah-medium-replay-v2 & 50.2 ± 0.2  & 34.3 ± 0.5  & 8.3 ± 0.6   & 2.1 ± 0.8   \\
halfcheetah-medium-v2        & 58.5 ± 0.2  & 45.2 ± 0.5  & 5.8 ± 0.1   & 1.5 ± 0.3   \\
walker2d-medium-expert-v2    & 111.1 ± 0.0 & 111.1 ± 0.0 & 110.8 ± 0.0 & 106.7 ± 0.9 \\
walker2d-medium-replay-v2    & 95.5 ± 0.5  & 91.1 ± 1.1  & 85.6 ± 1.4  & 45.2 ± 1.7  \\
walker2d-medium-v2           & 84.4 ± 0.0  & 84.8 ± 0.1  & 84.1 ± 0.1  & 80.3 ± 0.6  \\
hopper-medium-expert-v2      & 112.3 ± 0.1 & 111.4 ± 0.3 & 111.2 ± 0.3 & 110.3 ± 0.3 \\
hopper-medium-replay-v2      & 98.3 ± 0.9  & 87.5 ± 1.1  & 85.6 ± 4.8  & 90.3 ± 1.5  \\
hopper-medium-v2             & 84.6 ± 2.8  & 82.0 ± 1.6  & 83.5 ± 1.4  & 86.6 ± 0.6  
\end{tblr}
\caption{SAID's offline RL performance with error bar (S.E.M. 4 random seeds).}
\label{table:offline_results_supp}
\end{table}

\begin{table}
\centering
\begin{tblr}{
  cells = {c},
  cell{2}{1} = {r=4}{},
  cell{6}{1} = {r=4}{},
  cell{10}{1} = {r=4}{},
  cell{14}{1} = {r=4}{},
  hline{1,18} = {-}{0.08em},
  hline{2,6,10,14} = {-}{},
}
Task        & Delay & denoising step=2 & denoising step=5 & denoising step=10 & denoising step=20 \\
HalfCheetah & 0     & 14304.3 ± 272.1  & 14735.6 ± 350.1  & 14817.9 ± 362.5   & 14844.8 ± 366.7   \\
            & 4     & 9065.5 ± 425.4   & 9412.7 ± 446.3   & 9520.5 ± 455.9    & 9647.9 ± 435.5    \\
            & 8     & 5478.1 ± 278.6   & 5510.9 ± 269.5   & 5556.2 ± 278.3    & 5565.2 ± 273.7    \\
            & 16    & 5192.4 ± 413.2   & 5289.3 ± 425.9   & 5265.9 ± 424.3    & 5288.9 ± 431.8    \\
Ant         & 0     & 5586.9 ± 102.7   & 5900.8 ± 79.7    & 5964.3 ± 117.6    & 5953.4 ± 83.5     \\
            & 4     & 4540.8 ± 643.6   & 4615.8 ± 676.2   & 4544.5 ± 635.4    & 4585.6 ± 659.5    \\
            & 8     & 4436 ± 500.2     & 4507.6 ± 372.1   & 4507.3 ± 471.8    & 4609.9 ± 383      \\
            & 16    & 4171.4 ± 41.4    & 4134.9 ± 46.8    & 4123.8 ± 45.8     & 4137.4 ± 41.1     \\
Hopper      & 0     & 2750.7 ± 293.4   & 3086.7 ± 278.3   & 3145.6 ± 237.9    & 3170 ± 283.3      \\
            & 4     & 2856.9 ± 320.3   & 2798.8 ± 251.9   & 2744.6 ± 263      & 2697.7 ± 305.4    \\
            & 8     & 2916.1 ± 376.1   & 2805.2 ± 388.9   & 2537.4 ± 385.3    & 2513.7 ± 414.5    \\
            & 16    & 3321.7 ± 219.6   & 3260.5 ± 311.1   & 3261.8 ± 295      & 3268.1 ± 236.7    \\
Walker2d-v5    & 0     & 5002 ± 439.6     & 5122.1 ± 364.2   & 5147.1 ± 355.2    & 5158.1 ± 359.9    \\
            & 4     & 4805.1 ± 180.8   & 4851.5 ± 196.2   & 4874.2 ± 216.9    & 4874.1 ± 230.2    \\
            & 8     & 5108 ± 279.2     & 5182.7 ± 298.6   & 5210.5 ± 294.7    & 5222.4 ± 295.2    \\
            & 16    & 4814 ± 181.1     & 4916.1 ± 185.1   & 4921.6 ± 183.6    & 4913 ± 175.2      
\end{tblr}
\caption{SAID's online RL performance vs. Denoising steps of the diffusion model during inference.}
\label{table:denoising_steps}
\end{table}

\clearpage
\newpage

\newpage

\end{document}